\documentclass[sigconf]{aamas} 
\usepackage{balance} % for balancing columns on the final page

\usepackage{enumitem}
\usepackage{bm}
\usepackage{algorithm}
\usepackage[noend]{algorithmic}
\usepackage{caption}
\usepackage{subcaption}
\usepackage{multirow}

\setcopyright{ifaamas}
\acmConference[AAMAS '23]{Proc.\@ of the 22nd International Conference
on Autonomous Agents and Multiagent Systems (AAMAS 2023)}{May 29 -- June 2, 2023}
{London, United Kingdom}{A.~Ricci, W.~Yeoh, N.~Agmon, B.~An (eds.)}
\copyrightyear{2023}
\acmYear{2023}
\acmDOI{}
\acmPrice{}
\acmISBN{}

\acmSubmissionID{378}

\title[Diverse Policy Optimization for Structured Action Space]{Diverse Policy Optimization for Structured Action Space}

\author{Wenhao Li}
\affiliation{
  \institution{The Chinese University of Hong Kong, Shenzhen}
  \city{Shenzhen}
  \country{China}}
\email{liwenhao@cuhk.edu.cn}

\author{Baoxiang Wang}
\affiliation{
  \institution{The Chinese University of Hong Kong, Shenzhen}
  \city{Shenzhen}
  \country{China}}
\email{bxiangwang@cuhk.edu.cn}

\author{Shanchao Yang}
\affiliation{
  \institution{The Chinese University of Hong Kong, Shenzhen}
  \city{Shenzhen}
  \country{China}}
\email{shanchaoyang@link.cuhk.edu.cn}

\author{Hongyuan Zha}
\authornote{Corresponding author}
\affiliation{
  \institution{The Chinese University of Hong Kong, Shenzhen, Shenzhen Institute of AI and Robotics for Society}
  \city{Shenzhen}
  \country{China}}
\email{zhahy@cuhk.edu.cn}

\begin{abstract}
Enhancing the diversity of policies is beneficial for robustness, exploration, and transfer in reinforcement learning (RL).
In this paper, we aim to seek diverse policies in an under-explored setting, namely RL tasks with \textit{structured action spaces} with the two properties of \textit{composability} and \textit{local dependencies}.
The complex action structure, non-uniform reward landscape, and subtle hyperparameter tuning due to the properties of structured actions prevent existing approaches from scaling well.
We propose a simple and effective RL method, \textit{Diverse Policy Optimization (DPO)}, to model the policies in structured action space as the energy-based models (EBM) by following the probabilistic RL framework.
A recently proposed novel and powerful generative model, GFlowNet, is introduced as the efficient, diverse EBM-based policy sampler.
DPO follows a joint optimization framework: the outer layer uses the diverse policies sampled by the GFlowNet to update the EBM-based policies, which supports the GFlowNet training in the inner layer.
Experiments on ATSC and Battle benchmarks demonstrate that DPO can efficiently discover surprisingly diverse policies in challenging scenarios and substantially outperform existing state-of-the-art methods. 
\end{abstract}


\newcommand{\BibTeX}{\rm B\kern-.05em{\sc i\kern-.025em b}\kern-.08em\TeX}

\begin{document}

%%% The following commands remove the headers in your paper. For final 
%%% papers, these will be inserted during the pagination process.

\pagestyle{fancy}
\fancyhead{}

%%% The next command prints the information defined in the preamble.

\maketitle 

%%%%%%%%%%%%%%%%%%%%%%%%%%%%%%%%%%%%%%%%%%%%%%%%%%%%%%%%%%%%%%%%%%%%%%%%

\section{Introduction}\label{sec:intro}

The history of human civilization can be seen as a chronicle of creative capacity, i.e., the diversity of solutions to the same puzzle~\citep{osborn1953applied}.
Counter-intuitively, a popular consensus in deep learning with theoretical justifications~\citep{ma2021local} that most local optimas to a non-convex optimization problem are very close to the global optimum has led mainstream AI research to focus on finding a single local solution to a given optimization problem, rather than on which local optimum is dicovered~\citep{zhou2021continuously}.
It is no coincidence that most methods in reinforcement learning (RL) are also designed to seek a single reward-maximizing policy~\citep{sutton2018reinforcement,mnih2015human,schulman2017proximal}.

\begin{figure*}[htb!]
    \centering
    \includegraphics[width=0.8\textwidth]{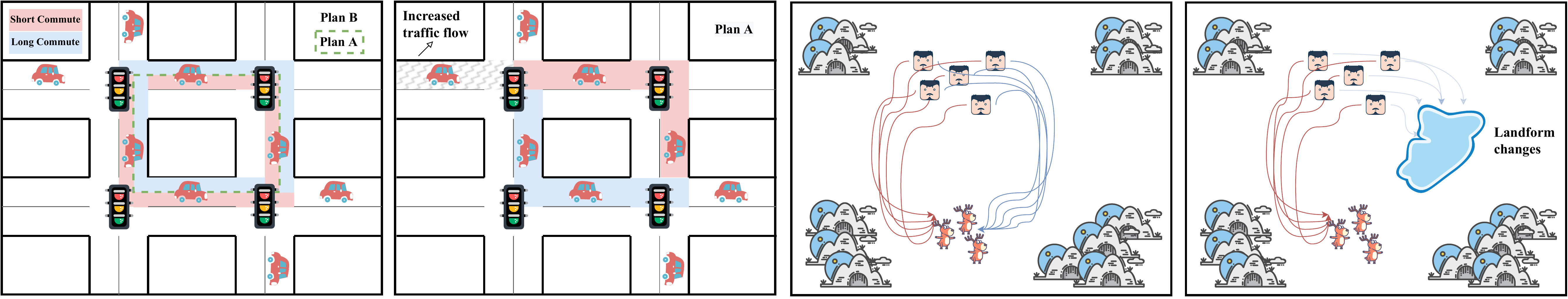}
    \caption{Robustness of diverse policies in two non-stationary environments: (Left) the adaptive traffic signal control and (Right) the predator-prey. In these tasks, diverse policies can quickly adapt to changes in the external environment.}
    \label{fig:illu}
\end{figure*}

However, different local optima in the policy space can correspond to strategies that differ in nature, which makes the above consensus problematic in RL tasks where the environment is unstable.
For example, in adaptive traffic signal control (ATSC)~\citep{van2016coordinated,wei2018intellilight,wei2019presslight} (conceptual diagram and more examples are included in Figure~\ref{fig:illu}), if two traffic flows are desired to reach the target points from the departure points quickly, multiple control strategies with similar average commuting times may exist due to the combinatorial nature of traffic lights.
The performance of a single policy obtained by reward maximization is bound to be affected if the subsequent traffic volumes on other sections of the road network associated with the traveled section of that traffic change.
Moreover, if our goal is to discover a diverse set of policies, some of these may prove more valuable than others in different situations.

Therefore, celebrating the diversity of policies is beneficial for many RL applications.
In addition to ATSC and the simple game in Figure~\ref{fig:illu}, these RL application areas include but are not limited to conversation generation in intelligent customer service~\citep{Li2016DeepRL}, drug discovery in smart healthcare~\citep{Pereira2021DiversityOD}, and simulator design in automated machine learning (AutoML)~\citep{Wang2019POETOC}.
Furthermore, in addition to robustness, a set of diverse policies can also be useful for exploration~\citep{peng2020non}, transfer~\citep{kumar2020one}, and hierarchy~\citep{alver2021constructing} in RL.

There is no doubt that RL researchers have demonstrated their creative ability in discovering diverse policies.
The majority of the literature has been done in the field of neuroevolution methods inspired by Quality-Diversity (QD), which typically maintains a collection of policies and adapts it using evolutionary algorithms to balance the QD trade-off~\citep{pugh2016quality,duarte2017evolution,parker2020effective,nilsson2021policy,gangwani2021harnessing,Lim2022DynamicsAwareQF}.
In another part of the work, intrinsic rewards have been used for learning diversity in terms of the discriminability of different trajectory-specific quantities~\citep{gregor2016variational,eysenbach2018diversity,hartikainen2019dynamical,goyal2019reinforcement,sharma2020emergent,zahavy2020planning,alver2021constructing}, or have been used as a regularizer when maximizing the extrinsic reward~\citep{Levine2018ReinforcementLA,gangwani2018learning,masood2019diversity,sharma2019dynamics,zhang2019learning}.
There is also a small body of work that transforms the problem into a Constrained Markov Decision Process (CMDP)~\citep{sun2020novel,zhou2021continuously,Derek2021AdaptableAP,Zahavy2022DiscoveringPW}, or implicitly induce diversity to learn policies that maximize the set robustness to the worst-possible reward~\citep{kumar2020one,Zahavy2021DiscoveringAS}.

This paper considers a more complex, realistic, less focused, and under-explored setting, namely RL tasks with \textit{structured action spaces}.
We define structured actions as actions with the following two properties:
\textit{composability}, i.e., environmental actions consist of a large number of atomic actions with complete functionality
% \footnote{Later on, the environmental action will be used uniformly to refer to the actions in the Markov decision process and the atomic action to refer to the sub-actions that make up the former.}; 
and \textit{local dependencies}, i.e., there are local physical or logical correlations between atomic actions\footnote{In this paper, only pairwise relationships between atomic actions are considered.}.
For example, in ATSC, the phases of all traffic signals on all intersections in the entire road network must be redetermined at certain intervals, and atomic actions are phases of each signal and interact with each other through the physical road network.
In addition, for the predator-prey task in Figure~\ref{fig:illu}, the atomic actions are the decisions of each predator, and there is a local spatial, logical association.

The high dimensionality of the RL agent's policy due to the composability of structured actions prevents existing methods from scaling well.
Specifically, the combinability will make the underlying reward landscape of the RL problem particularly non-uniform, which may make QD-like methods require substantially large population sizes to fully explore the policy space and prevent the algorithm from collapsing to visually identical policies~\citep{tang2020discovering,zhou2021continuously}.
Also, due to composability, the complex soft objective introduced by intrinsic reward or CMDP-driven methods will result in non-trivial and subtle hyperparameter tuning~\citep{masood2019diversity,parker2020effective}.
In addition, the existing agents' policies are mainly parameterized categorical distributions or Gaussian distributions. 
Their extension to structured actions with independent assumptions on atomic actions will prevent the agent from effectively using the structural information of environmental actions to achieve an efficient search for the policy space.

We propose a simple and effective RL method, \textit{Diverse Policy Optimization (DPO)}, to discover a diverse set of policies in tasks with structured action spaces.
We follow the probabilistic reinforcement learning (PRL) framework~\citep{Levine2018ReinforcementLA} to transform reinforcement learning problems under stochastic dynamics into variational inference problems on probabilistic graphical models and model the policies of RL agents as the energy-based models (EBM).
The action distribution induced by this EBM in a structured action space is highly multimodal, and sampling from such a high-dimensional distribution is intractable.
To this end, we introduce a recently proposed novel and powerful generative model, \textit{Generative Flow Networks (GFlowNet)}~\citep{bengio2021flow,bengio2021gflownet,jain2022biological,zhang2022generative}, as the efficient diverse policy sampler.
GFlowNet can be regarded as amortized Monte-Carlo Markov chains (MCMC), which gradually builds composable environmental actions through the single but trained generative pass of "building blocks (i.e., atomic actions)", so that the final sampled environmental actions obey a given energy-based policy distribution.

Notably, our method does not simply introduce the GFlowNet to RL with structured action spaces.
Since in the PRL framework, with the update of the soft Q function, the energy-based policy distribution is also constantly changing.
This violates the assumption of the fixed energy model in GFlowNet and makes DPO face a more complex optimization problem.
Therefore, we model DPO as a joint optimization problem: the outer layer uses the diverse policies sampled by the GFlowNet to update the soft Q function, and the inner layer trains the GFlowNet through an EBM based on the soft Q function (see Figure~\ref{fig:framework}).
Furthermore, a two-timescale alternating optimization method is proposed to solve it efficiently.

We empirically validate DPO on ATSC tasks~\citep{ault2021reinforcement} where atomic actions have local physical dependencies, and more generally, Battle scenarios~\citep{zheng2018magent} where atomic actions have logical local dependencies.
Experiments demonstrate that DPO can reliably and efficiently discover surprisingly diverse strategies in all these challenging scenarios and substantially outperform existing baselines. 
The contributions can be summarized as follows:
\begin{enumerate}[label=(\arabic*),leftmargin=13pt]
    \item We propose a novel algorithm, \textit{Diverse Policy Optimization}, for discovering diverse policies for structured action spaces. The GFlowNet-based sampler can efficiently sample diverse policies from the high-dimensional multimodal distribution induced by structured action spaces.
    \item We propose an efficient joint training framework to interleaved optimize the soft-Q-function-based EBM and the reward-conditional GFlowNet-based sampler.
    \item Our algorithm is general and effective across structured action spaces with physical and logical local dependencies.
\end{enumerate}

%%%%%%%%%%%%%%%%%%%%%%%%%%%%%%%%%%%%%%%%%%%%%%%%%%%%%%%%%%%%%%%%%%%%%%%%

\section{Preliminaries and Notations}\label{sec:pre}

The proposed DPO follows the PRL to model policies as a high-dimensional multimodal energy-based probability distribution and introduces GFlowNet to efficiently sample policies with diversity from this distribution. 
Below, we briefly review the PRL and GFlowNet.

\subsection{Probabilistic Reinforcement Learning}\label{sec:prl}

PRL aims to learn the maximum entropy optimal policy:  
\begin{equation}
    \pi_{\mathrm{ent}}^{*}=\arg \max _{\pi} \sum_{t} \mathbb{E}_{\left({s}^{t}_{e}, a^{t}_{e}\right) \sim \rho_{\pi}}\left[r\left({s}^{t}_{e}, {a}^{t}_{e}\right)+\alpha \mathcal{H}\left(\pi\left(\cdot \mid {s}^{t}_{e}\right)\right)\right], \nonumber
\end{equation}
where $s^{t}_{e}\in\mathcal{S}_{e}$ and $a^{t}_{e}\in\mathcal{A}_{e}$ denotes the state and action respectively.
The subscript $e$ represents the ``environment'', which is used to distinguish related concepts in RL from GFlowNets, and the $\alpha$ is the coefficient to trade off between entropy and reward.
Function $\mathcal{H}$ denotes the entropy term.
By defining the soft $Q$ function as:
\begin{equation}
    \resizebox{0.9\hsize}{!}{$
        Q_{\mathrm{soft}}^{*} \!\left({s}^{t}_{e}, {a}^{t}_{e}\right)\!:=r^{t}_{e}+ \mathbb{E}_{{s}^{t+\ell}_{e} \sim \rho_{\pi}}\left[\sum_{\ell=1}^{\infty} \gamma^{\ell}\left(r^{t+\ell}_{e}+\alpha \mathcal{H}\left(\pi_{\mathrm{ent}}^{*}\left(\cdot \mid {s}^{t+\ell}_{e}\right)\right)\right)\right].$
    }
\end{equation}
The optimal maximum entropy policy can be proved as in \cite{Levine2018ReinforcementLA}
\begin{equation}
    \pi_{\mathrm{ent}}^{*} = \exp\left({\frac{1}{\alpha}\left(Q_{\mathrm{soft}}^{*}\left({s}^{t}_{e}, {a}^{t}_{e}\right)-V_{\mathrm{soft}}^{*}\left({s}^{t}_{e}\right)\right)}\right)\, ,\label{eq:energy-based-policy}
\end{equation} where the soft value function $V_{\mathrm{soft}}^{*}$ is defined by
\begin{equation}
    V_{\mathrm{soft}}^*\left(s^t_{e}\right)=\alpha \log \int_{\mathcal{A}_{e}} \exp \left(\frac{1}{\alpha} Q_{\mathrm{soft}}^*\left(s^t_{e}, a^{\prime}_{e}\right)\right) d a^{\prime}_{e}.
\end{equation}
Thus the policy learning can be treated as the approximation to the Boltzmann-like distribution of optimal $Q$ function.
Taking the soft $Q$-Learning (SQL)~\cite{sql} method as an example, it provides the optimal $Q$ is the fixed point of soft Bellman backup, which satisfies the soft Bellman equation
\begin{equation}
    Q_{\mathrm{soft}}^*\left(s^t_{e}, a^t_{e}\right)=r^t_{e}+\gamma \mathbb{E}_{s^{t+1}_{e} \sim p_{s_{e}}}\left[V_{\mathrm{soft}}^*\left(s^{t+1}_{e}\right)\right].
\end{equation}
Due to the infinite set of states and actions, it takes parameterized $Q$ and uses a function $\pi$ as an approximate sampler of Boltzmann-like distribution of $Q$.
Specifically, it updates $Q$ and $\pi$ as:
\begin{equation}
    \left\{
        \begin{aligned}
            &\resizebox{0.8\hsize}{!}{$\min_{\theta} J_{Q}(\theta) := \mathbb{E}_{s^{t}_{e}, a^{t}_{e}, r^{t}_{e}, s^{t+1}_{e} \sim D}\left[\frac{1}{2}\left(r^{t}_{e} + V^{\bar{\theta}}\left(s^{t+1}_{e}\right)-Q^{\theta}\left(s^{t}_{e}, {a}^{t}_{e}\right)\right)^{2}\right],$}\\
            &\resizebox{0.8\hsize}{!}{$\min_{\phi} J_{\pi}\left(\phi ; s^{t}_{e}\right) :=  {\mathrm{KL}}\left(\pi^{\phi}\left(\cdot | s^{t}_{e}\right) \| \exp \left(\frac{1}{\alpha}\left(Q^{\theta}\left(s^{t}_{e}, \cdot\right)-V^{{\theta}}(s^{t}_{e})\right)\right)\right),$}
            \label{eq: sql0-policy}
        \end{aligned}
    \right.
\end{equation}
where function $V^{{\theta}}$ is denoted as
\begin{equation}\label{eq:soft-value}
    V^{{\theta}}\left(s^{t}_{e}\right) :=  \alpha \log \mathbb{E}_{{a}^{\prime}_{e}\sim q_{a^{\prime}_{e}}}\left[\exp \left(\frac{1}{\alpha} Q^{{\theta}}\left(s^{t}_{e}, {a}^{\prime}_{e}\right)\right)/q_{{a}^{\prime}_{e}}({a}^{\prime}_{e})\right],
\end{equation}
and $\theta, \bar{\theta}, \phi$ denote the parameters of critic, target critic and policy respectively;
$q_{a^\prime}$ is an arbitrary policy distribution.
The policy distribution induced by the EBM (i.e., the Boltzmann-like distribution of $Q$) under structured action spaces is highly multimodal, and sampling from such a high-dimensional distribution is intractable.
In this paper, DPO introduces a powerful generative model, \textit{Generative Flow Networks (GFlowNet)}, as the efficient diverse policies sampler.

\subsection{Generative Flow Networks}\label{sec:gfn}

\begin{figure*}[htb!]
    \centering
    \includegraphics[width=0.8\textwidth]{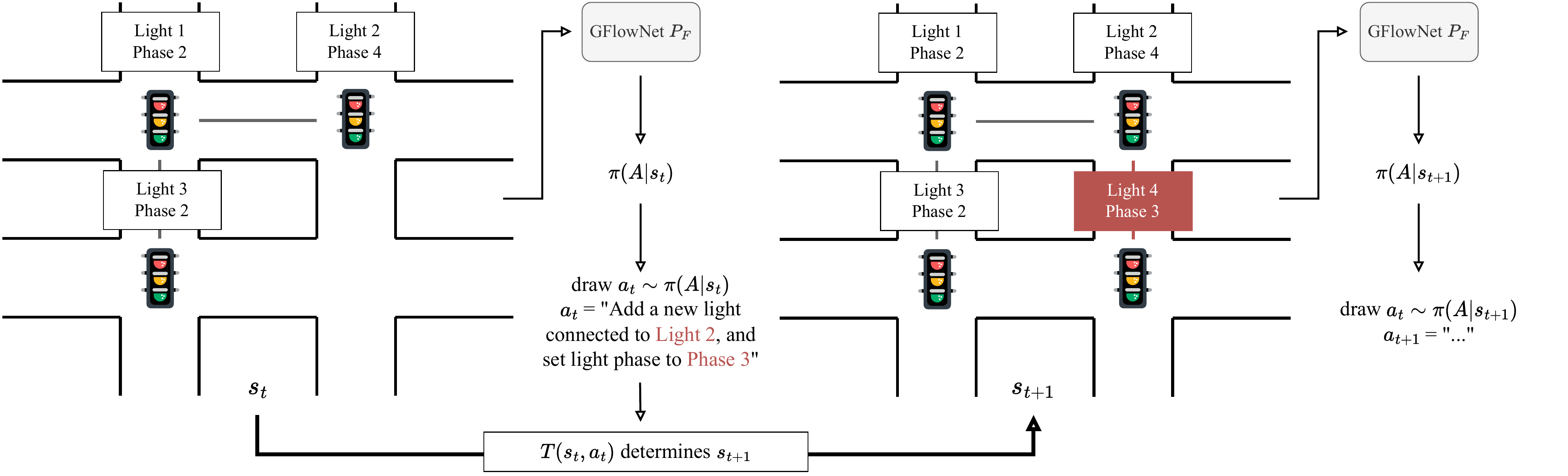}
    \caption{A GFlowNet iteratively constructs an composite object, e.g., a traffic light network. $s_t$ represents the state of the partially constructed object, $a_t$ represents the action taken by the GFLowNet to transition to state $s_{t+1}=T(s_t,a_t)$. The GFlowNet take a 3-lights traffic network as input and determines an action to take. This process repeats until an exit action is sampled or maximum light number is achieved and the sample is complete.}
    \label{fig:gfn}
\end{figure*}

Generative flow networks, which are trainable generative policies, model the generation or sampling process of composite objects $x \in \mathcal{X}$  by a sequence of discrete \textit{actions} that incrementally modify a partially constructed object (\textit{state}).
Note that action and state here do not refer to the concepts in RL~\citep{zhang2022generative}.
In this paper, we model \textit{actions in RL} problems with structured action spaces as \textit{states in GFlowNet}, and \textit{actions in GFlowNet} correspond to the \textit{atomic actions} that compose structured actions.
In other words, the composite object $x$ generated by the GFlowNet is $a_e$, and $\mathcal{X}$ is equivalent to $\mathcal{A}_{e}$.
The partially constructed object and corresponding action sequence space can be represented by a directed acyclic graph~(DAG, See the DAG consisting of traffic lights and roads in Figure ~\ref{fig:gfn})~$G=(\mathcal{S}_{g}, \mathcal{A}_{g})$, where the subscript $g$ denotes the ``GFlowNet''.
The vertices in $\mathcal{S}_{g}$ are states and the edges in $\mathcal{A}_{g}$ are actions that modify one state to another.
The tails of incoming edges and the heads of outgoing edges of a state are denoted as the \textit{parents} and \textit{childrens}, respectively.
The sampling process of the composite object $a_e$ starts from the \textit{initial state} $s^0_{g}$ and transits to the \textit{terminal} state $s^n_{g} \in \mathcal{A}_{e}$, which is a state without outgoing edges, after $n \in (0, T]$ steps and $T$ is the maximum length.
Note that the same terminal state may correspond to multiple action sequences.

A \textit{complete trajectory} is a state sequence from a initial state to a terimal state $s^0_{g} \rightarrow s^1_{g} \rightarrow \ldots \rightarrow s^n_{g}$, where each transition $s^t_{g} \rightarrow s^{t+1}_{g}$ is an action in $\mathcal{A}_{g}$.
A \textit{trajectory flow} is a unnormalized density or a non-negative function, $F: \mathcal{T} \rightarrow \mathbb{R}_{\geq 0}$, on the set of all complete trajectories $\mathcal{T}$.
The flow is called \textit{Markovian} if there exist distributions $P_F(\cdot \mid s_{g})$ over the children of every non-terminal state $s_{g}$ and a constant $Z$, such that for any complete trajectory $\tau$ we have $P_F(\tau)=F(\tau) / Z$ with $P_F(\tau)=P_F\left(s^1_{g} \mid s^0_{g}\right) P_F\left(s^2_{g} \mid s^1_{g}\right) \ldots P_F\left(s^n_{g} \mid s^{n-1}_{g}\right)$.
$P_F\left(s^{t+1}_{g} \mid s^t_{g}\right)$ is called a \textit{forward policy}, which is used to sample the composite object $a_e$ from the density $F$.
$P_T(a_e)$ then denotes the probability that a complete trajectory sampled from $P_F$ terminates in $a_e$.

The problem we are interested in is fitting a Markovian flow to a fixed \textit{energy function} on $\mathcal{A}_{e}$.
Given an energy function $\mathcal{E}(a_e):=-\log R(a_e)$ and the associated non-negative \textit{reward function} (again, not a reward in RL) $R_{g}: \mathcal{A}_{e} \rightarrow \mathbb{R}_{\geq 0}$, one seeks a Markovian flow $F$ such that the likelihood of a complete trajectory sampled from $F$ terminating in a given $a_e$ is proportional to $R_{g}(a_e)$, i.e., $P_T(a_e) \propto R_{g}(a_e)$.
This $F$ can be obtained by imposing the \textit{reward-matching} constraint: $R_{g}(a_e)=\sum_{\tau=\left(s^0_{g} \rightarrow \ldots \rightarrow s^n_{g}\right), s^n_{g}=a_e} F(\tau)$.
The details of how to parameterize a GFlowNet and train a Markovian flow $F$ that satisfies the reward matching constraint will be explained soon.

%%%%%%%%%%%%%%%%%%%%%%%%%%%%%%%%%%%%%%%%%%%%%%%%%%%%%%%%%%%%%%%%%%%%%%%%

\section{Diverse Policy Optimization}\label{sec:dpo}

This section proposes a simple and effective RL method, \textit{Diverse Policy Optimization (DPO)}, to discover diverse policies in structured action spaces.
We follow the probabilistic reinforcement learning (PRL) framework~\citep{Levine2018ReinforcementLA} to transform RL problems under stochastic dynamics into variational inference problems on probabilistic graphical models and model the policies of RL agents as EBMs.
PRL framework corresponds to a maximum entropy variant of reinforcement learning or optimal control, where the optimal policy aims to maximize the expected reward and maintain high entropy.
Due to the maximum entropy objective, some existing works~\citep{sql,haarnoja2018soft} have proposed algorithms for low-dimensional continuous action spaces to discover diverse policies based on this framework.

Our method is an instance of the maximum entropy actor-critic algorithm in the PRL framework, which adopts a message-passing approach and can produce lower-variance estimates.
In addition, to make the policy still scalable in the structured action space, we do not use an explicit policy parameterization but fit only the message, i.e., the $Q$-value function, similar to soft $Q$-learning~\citep{sql}.
Specifically, we opt for using general energy-based policies $\pi\left(a_{e} \mid s_{e}\right) \propto \exp \left(-\mathcal{E}\left(s_{e}, a_{e}\right)\right),$
% \begin{equation}
    % \pi\left(a_{e} \mid s_{e}\right) \propto \exp \left(-\mathcal{E}\left(s_{e}, a_{e}\right)\right),
% \end{equation} 
where $\mathcal{E}$ is an energy function.
Furthermore, we set $\mathcal{E}\left(s_{e}, a_{e}\right)=-\frac{1}{\alpha} Q_{\text {soft }}\left(s_{e}, a_{e}\right)$, then the optimal maximum entropy policy is an EBM that satisfies Equation~(\ref{eq:energy-based-policy}).

However, The action distribution induced by this EBM in a structured action space is highly multimodal, and sampling from such a high-dimensional distribution is intractable.
Fortunately, the composability and local dependencies of the structured action space make generative flow networks naturally suitable for efficiently sampling diverse and high-quality policies from it.
And we only need to set the energy function that needs to be fitted by the Markovian flow $F(a_{e})$ (where the action $a_{e}$ corresponding to the composite object $x$) to be $(-{1}/{\alpha})\cdot Q_{\text {soft }}\left(s_{e}, a_{e}\right)$, and its associated reward function $R_{g}(a_{e})$ to be set to $\exp\left(({1}/{\alpha})\cdot Q_{\text {soft }}\left(s_{e}, a_{e}\right)\right)$, we can elegantly introduce GFlowNet as an efficient and diverse sampler.

Nevertheless, the unreasonable part of the above modeling is that there is no place left for the environment state $s_{e}$ in the input of the Markovian flow and the reward function.
The reason is that $\pi$ in the PRL framework is a \textit{conditional} distribution, but GFlowNet is an \textit{unconditional} sampler. 
To this end, we will introduce a variant of GFlowNet, namely \textit{reward-conditional GFlowNet}, to model the policy of RL agents, and details will be explained shortly.

\begin{figure}[htb!]
    \centering
    \includegraphics[width=0.4\textwidth]{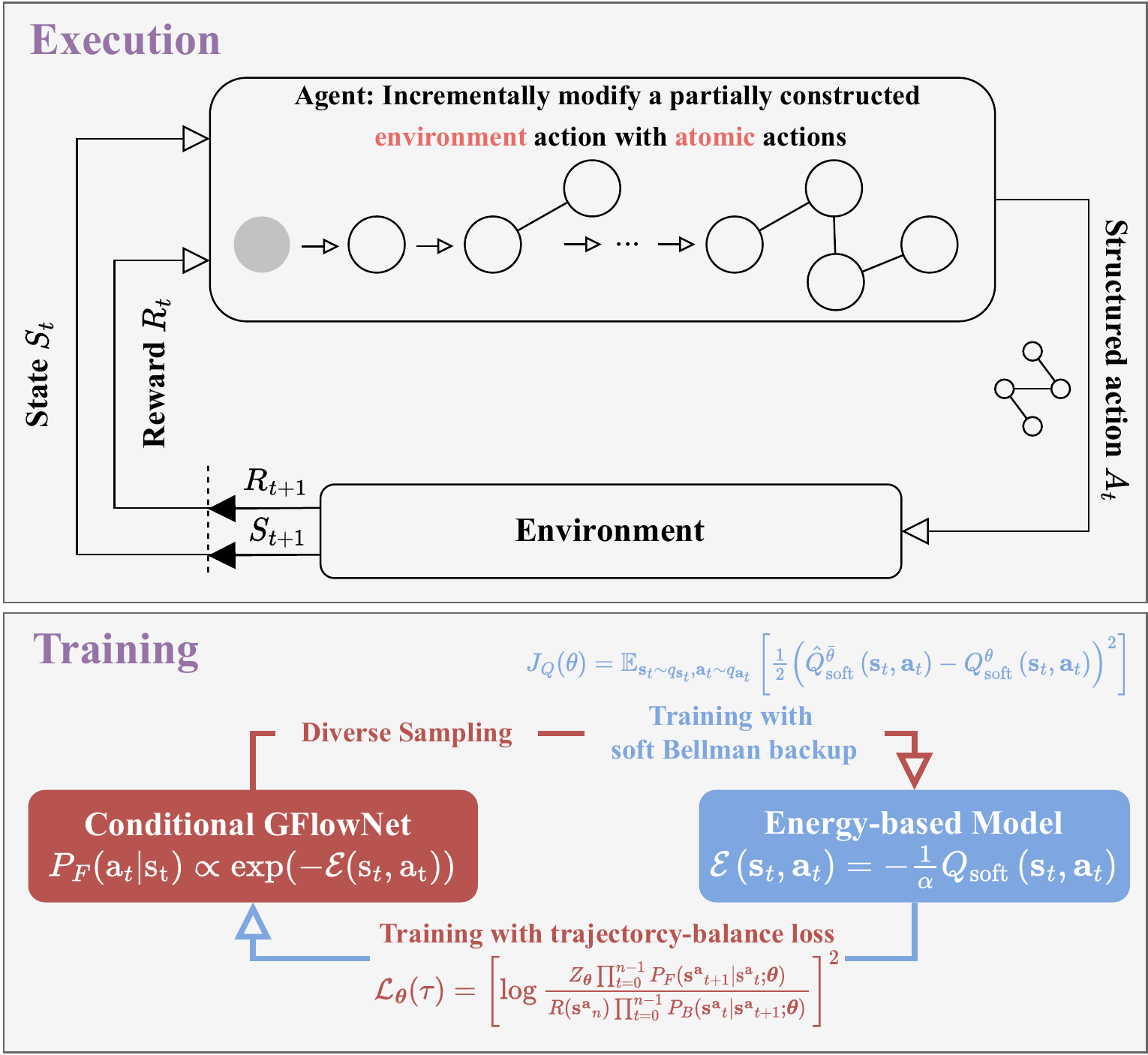}
    \caption{The schematic diagram of RL based on GFlowNet as the agent's policy and the joint training framework.}
    \label{fig:framework}
\end{figure}

Since in the PRL framework, with the update of the $Q_{\mathrm{soft}}$, the energy-based policy distribution is also constantly changing.
DPO adopts a joint training framework where the EBM and the GFlowNet are optimized alternately, similar with~\citep{zhang2022generative}:
The energy function serves as the negative log-reward function for the GFlowNet, which is trained with the trajectory balance~\citep{malkin2022trajectory} objective to sample from the evolving energy-based policies. 
In contrast, the energy function is trained with soft Bellman backup, where the GFlowNet provides diverse samples.
The schematic diagram of RL based on reward-conditional GFlowNet as the agent's policy and the joint training framework are shown in Figure~\ref{fig:framework} and Algorithm~\ref{alg:dpo}.

In the following, we will explain the generation process of structured action, the parameterization and training of reward-conditional GFlowNet, and its interleaved update with EBM, respectively.

\subsection{Structured Action Generative Process}\label{sec:act-gen}

The framework of diverse policy optimization is introduced in the previous section, and this section will describe the process of generating structured actions based on the reward-conditional GFlowNet.
The local dependencies of structured actions indicate that there may be two correlations between atomic actions: locally physical and locally logical correlations. 
The former is a typical graph, while the latter belongs to a typical set.
For the unity of the framework, this paper only considers the physical correlation between atomic actions. 
It transforms the logical correlation into the physical correlation without loss of generality.

Expressly, we assume that atomic actions with local logical correlations have a fixed influence range with a radius $d$ in Euclidean space. 
An atomic action can then establish a physical correlation with others within its influence range.
Of course, other types of topologies, such as fully connected, star, hierarchical, etc., can also be used in addition to adjacency topologies. 
This paper adopts the adjacency topology to make a trade-off between efficiency and performance. 
The experimental results also show that the algorithm performance is not sensitive to the influence radius $d$.

In the structured action space, the action consists of $N$ atomic actions in $K$-dimensional discrete space, i.e., $a_e \in \mathcal{A}_e \triangleq [K]^{N}$, where $[K] \triangleq \{0, \ldots, K-1\}$.
$a_e$ could be a phase configuration of $N$ traffic lights, and each traffic light contains $K$ phases or the joint action of $N$ predators, and each predator can go in $K$ directions. 
We model the generation or sampling of vectors in $\mathcal{A}_e$ by a reward-conditional GFlowNet.
The state space of GFlowNet is denoted as $\mathcal{S}_{g}$,
% to distinguish it from the state space $\mathcal{S}_{e}$ in RL,
and we have $\mathcal{S}_{g} \triangleq\left\{\left(s_g^1, \ldots, s_g^N\right) \mid s_g^n \in [K] \cup \oslash, n=1, \ldots, N\right\},$
% \begin{equation}
%     \mathcal{S}_{g} \triangleq\left\{\left(s_g^1, \ldots, s_g^N\right) \mid s_g^n \in [K] \cup \oslash, n=1, \ldots, N\right\},
% \end{equation}
where the void symbol $\oslash$ represents a yet unspecified atomic action.
The DAG structure on $\mathcal{S}_{g}$ is the $N$-th Cartesian power of the DAG with states $[K] \cup \oslash$, where $[K]$ are children of $\oslash$.
Concretely, the children of a state $s_g=\left(s_g^1, \ldots, s_g^N\right)$ are vectors that can be obtained from $s_g$ by changing any one atomic action $\mathbf{s}_{\mathrm{g}}^n$ from $\oslash$ to $[K]$, and its parents are states that can be obtained by changing a single atomic action $s_g^n \in [K]$ to $\oslash$.

Moreover, $\mathcal{A}_e$ is naturally identified with $\{s_g \in \mathcal{S}_{g}:|s_g|=D\}$ where $|s_g| \triangleq \#\left\{s_g^n \mid s_g^n \in [K], n=1, \ldots, N\right\}$.
Similarly, the initial state is denoted as $s^{0}_g \triangleq$ $(\oslash, \oslash, \ldots, \oslash)$, which means that the reward-conditional GFlowNet-based RL policy needs to take $N$ steps to sample a structured action, i.e., constructing a trajectory from $s^{0}_g$ to $a_e \in \mathcal{A}_{e}$.
The forward policy $P_{F|e}(\cdot|s_g, s_e)$ of a reward-conditional GFlowNet (will explained soon), extends from $\S$\ref{sec:gfn}, is a distribution over all paths to select a position with a void atomic action in $s_g$ and a value $k \in [K]$ to assign to this atomic action based on the environmental state $s_e \in \mathcal{S}_{e}$. 
Thus the action space for a state $s_g$ has size $K(N-|s_g|)$.
Since $k \ll N$, the action space of the forward policy (same as the backward policy below) grows \textit{linearly} with the atomic actions increase, so DPO has a good scalability.
Correspondingly, the backward policy $P_{B|e}(\cdot|s_g, s_e)$ is a distribution over the $|s_g|$ paths to select a position with a nonvoid atomic action in $s_g$.\\

\begin{figure*}[htb!]
    \centering
    \includegraphics[width=0.8\textwidth]{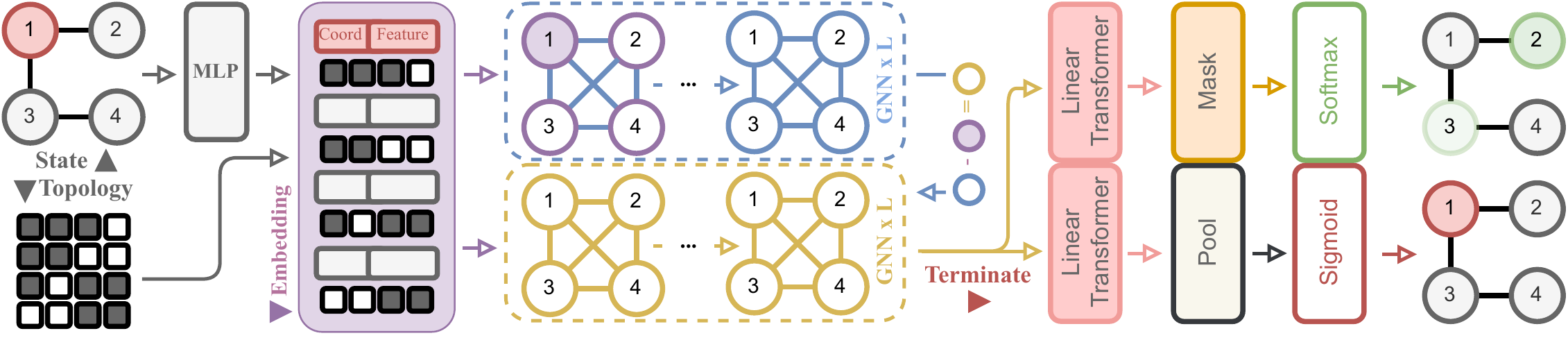}
    \caption{The parameterized forward and backward policy based on the modified graph pointer network.}
    \label{fig:gpn}
\end{figure*}

\noindent\textbf{More efficient generation.}
As we mentioned earlier, as an amortized version of MCMC, GFlowNets can alleviate the mix-moding problem~\citep{jasra2005markov,pompe2020framework} of the MCMC method, thereby improving the sampling efficiency of diverse samples.
However, if the two modes are close enough, the MCMC method will have higher sampling efficiency because it only perturbs the previous sample slightly. 
However, GFlowNets, for this case, need to rebuild the entire structured action sequentially, although only a minimal number of atomic actions have changed.
To this end, we introduce a small trick: adding a \textit{termination} action in the action space. 
GFlowNets are trained to successfully sample from two close modes by deciding to terminate at different modes at different runs.
% corresponding to each state to terminate the construction process of the structured action, and the remaining unprocessed atomic actions are consistent with the last generation.
Since the physical meaning of the termination action is quite different from other actions, we use a different output head to predict it separately, as shown in Figure~\ref{fig:gpn}. 
Once the forward policy $P_{F|e}(\cdot|s_g, s_e)$ decides to take the termination action, the output of the other head will be ignored.
Experiments show that this small trick can significantly improve the learning efficiency of the algorithm in some tasks.

\subsection{GFlowNet Parameterization}\label{sec:net-param}

After showing how to sample structured actions using the GFlowNet, this section elaborates on how to parameterize it and train a Markovian flow $F$ that satisfies the reward matching constraint.
As stated earlier, if we take the form of the GFlowNet in $\S$\ref{sec:gfn}, there will be no place for the environment state $s_e$ in the forward policy $P_F$ as well as in the backward policy $P_B$.
Thus, we use an extended version of flow networks by conditioning each component on some information, which is \textit{external} to the flow network but influences the terminating flows.
In our setting, the external information is RL's environmental state $s_e$.
Since the external information $s_e$ affects the reward function $R_g$ in $\S$\ref{sec:gfn}, this conditional GFlowNet is also called \textit{reward-conditional GFlowNet}~\citep[Definition 29]{bengio2021gflownet}.

Since reward-conditional GFlowNets are defined using the same components as the unconditional one, they inherit from all the properties of the GFlowNet for all DAGs $G_{e}=(\mathcal{S}_g, \mathcal{A}_g, \mathcal{S}_e)$ and flow functions $F_{e}: \mathcal{T} \times \mathcal{S}_e \rightarrow \mathbb{R}_{\geq 0}$, where $e$ represents the ``environment'' in RL again.
In particular, we can directly extend notions of $\S$\ref{sec:gfn} to reward-conditional GFlowNets with forward policy $P_{F|e}(\cdot|s_g, s_e)$, backward policy $P_{B|e}(\cdot|s_g, s_e)$, energy function $\mathcal{E}(a_e|e):=-\log R_(g|e)(a_e|s_e)$ and the associated non-negative reward function $R_{g|e}: \mathcal{A}_e \times \mathcal{S}_{e} \rightarrow \mathbb{R}_{\geq 0}$;
The only difference is that now every term explicitly depends of the conditioning variable, environmental state $s_e \in \mathcal{S}_{e}$ under the RL context.

In our experiments, we parameterize the forward and backward policy with deep neural networks $P_{F|e}(\theta_F)$ and $P_{B|e}(\theta_B)$ respectively, and for convenience, we omit the input without introducing ambiguity.
As $P_F$ incrementally builds structured actions, its action space gradually decreases, similar to the traveling salesman problem (TSP)~\citep{tsp}.
Considering the effectiveness of the pointer network~\citep{ptn} in dealing with such problems, we introduce the modified graph pointer network (GPN,~\citep{gpn}) as the forward and backward policy (see Figure~\ref{fig:gfn}) to further model the structured information of the action space.
The forward process of the modified GPN can be divided into the following three stages:

\noindent\textbf{Environmental state encoding: }
In this stage, the $i$-th row of the adjency matrix $\ell_i$ and local observed information $o_i$ of each atomic action are concatenated as $s_{i|e}=\left[\ell_i \| o_i\right]$, and then $s_{i|e}$ is embedded into a higher dimensional vector $\tilde{s}_{i|e} \in \mathbb{R}^d$ by a shared feed-forward network, where $d$ is the hidden dimension.
The context information is then obtained by encoding all atomic actions' embeddings $s_e$ via a graph neural network (GNN,~\citep{kipf2016semi,xu2018powerful}), where $s_e=[\tilde{s}_{1|e}^{\top}, \ldots, \tilde{s}_{N|e}^{\top}]^{\top}$.
Each layer of the GNN is expressed as:
\begin{equation}
    s_{i|e}^{\ell}=\gamma s_{i|e}^{\ell-1} \Theta+(1-\gamma) \phi_\theta\left(\frac{1}{|\mathcal{N}(i)|}\left\{s_{j|e}^{\ell-1}\right\}_{j \in \mathcal{N}(i) \cup\{i\}}\right),
\end{equation} where $s_{i|e}^\ell \in \mathbb{R}^{d_{\ell}}$ is the $\ell$-th layer variable with $\ell \in\{1, \ldots, L\}$, $s_{i|e}^0=s_{i|e}$, $\gamma$ is a trainable parameter,
% which regularizes the eigenvalue of the weight matrix,
$\Theta \in \mathbb{R}^{d_{\ell-1} \times d_\ell}$ is a trainable weight matrix, $\mathcal{N}(i)$ is the adjacency set of atomic action $i$, and $\phi_\theta: \mathbb{R}^{d_{\ell-1}} \rightarrow \mathbb{R}^{d_\ell}$ is the aggregation function~\citep{kipf2016semi}, which is represented by a neural network in our experiments.

\noindent\textbf{GFlowNet state encoding: }
In this stage, we use the vectors pointing from the newly added atomic action to all others as the embedding of $s_g$, which is similar with~\citet{gpn}.
Specifically, for the newly added atomic action $\tilde{s}_{i|e}$, suppose $s_{\mathrm{E|i}}=\left[\tilde{s}_{i|e}^{\top}, \ldots, \tilde{s}_{i|e}^{\top}\right]^{\top} \in \mathbb{R}^{N \times d}$ is a matrix with identical rows $\tilde{s}_{1|e}$. 
We define $s_g=s_{i|e}^L-s_{\mathrm{E|i}}=\left[s_{i|g}^{\top}, \ldots, s_{N|g}^{\top}\right]^{\top} \in \mathbb{R}^{N \times d}$.
Then $s_{g}$ is passed into the GNN again and the embedding of each atomic action after GFlowNet state encoding is denoted as $s_{i|g}^L$.

\noindent\textbf{Atomic action selection: }
The atomic action selector is based on the Linear Transformer~\citep{katharopoulos2020transformers}, which has the advantage of not suffering from the quadratic scaling in the input size.
This architecture relies on a linearized attention mechanism, defined as
\begin{equation}
    \begin{gathered}
        Q=s_{g}^L W_Q \quad K=s_{g}^L W_K \quad V=s_{g}^L W_V, \\
        \operatorname{LinAttn}_k(s_{g}^L)=\frac{\sum_{j=1}^N\left(\psi\left(Q_k\right)^{\top} \psi\left(K_j\right)\right) V_j}{\sum_{j=1}^N \psi\left(Q_k\right)^{\top} \psi\left(K_j\right)},
    \end{gathered}
\end{equation} where $\psi(\cdot)$ is a non-linear feature map, and $Q, K$, and $V$ are linear transformations of $s_{g}^L$ corresponding to the queries, keys, and values respectively, as is standard with Transformers.
The pointer vector outputted by the Linear Transformer is first masked by the mask $\mathbf{m}$ associated with the physical dependencies in structured action space and is then passed to a softmax layer to generate a distribution over the next candidate intersections.
Similar to pointer networks~\citep{ptn}, the masked pointer vector $\mathbf{u}_{i}$ is defined as:
\begin{equation}
\mathbf{u}_{i}^{(j)}= \begin{cases}\mathbf{u}_{i}^{(j)} & \text { if } j \neq \sigma(k), \forall k<j, \\ -\infty & \text { otherwise, }\end{cases}
\end{equation}
where $\sigma(k)$ denotes $k$-th processed atomic action and $\mathbf{u}_{i}^{(j)}$ is the $j$-th entry of the vector $\mathbf{u}_{i}$.

\subsection{Reward-Conditional GFlowNet Training}

After parameterizing the GFlowNet, we now describe how reward-conditional GFlowNets could be trained toward matching a given conditional reward.
Recall from $\S$\ref{sec:gfn}, $\S$\ref{sec:dpo} and $\S$\ref{sec:act-gen}, given a non-negtive conditional reward function $R_{g|e}: \mathcal{A}_{e} \times \mathcal{S}_{e} \rightarrow \mathbb{R}_{\geq 0}$, a reward-conditional GFlowNet can be trained so that its terminating probability distribution matches the associated energy-based model.
To be precise, the marginal likelihood that a trajectory sampled from the forward policy $P_{F|e}(\cdot|s_g, s_e)$ terminates at a given structured action is propotional to the action's soft $Q$ value $P_T(a_e|s_e) \propto \exp\left(({1}/{\alpha})\cdot Q_{\text {soft }}\left(s_e, a_e\right)\right)$, where $a_e \in \mathcal{A}_{e}$ and $s_e \in \mathcal{S}_{e}$.

To train the parameters $\theta_F$ and $\theta_B$ of the reward-conditional GFlowNet, we use the trajectory balance objective~\citep{malkin2022trajectory} that optimizes the following objective along complete trajectories $\tau=(s_g^0 \rightarrow s_g^1 \rightarrow \ldots \rightarrow \ldots \rightarrow s_g^n)$:
\begin{equation}\label{eq:tbloss}
    \mathcal{L}_{\Theta}(\tau | s_e)=\left[\log \frac{Z\left(s_e;\theta_Z\right) \prod_{t=0}^{n-1} P_F\left(s_g^{t+1} | s_g^{t}, s_e ; \theta_F\right)}{R\left(s_g^n | s_e\right) \prod_{t=0}^{n-1} P_B\left(s_g^{t} | s_g^{t+1}, s_e ; \theta_B \right)}\right]^2,
\end{equation} where $\Theta \triangleq \{\theta_F, \theta_B, \theta_Z\}$.
The scalar function $Z(\cdot)$ is parametrized in the log domain, as suggested by~\citet{malkin2022trajectory}.
With the trajectory balance objective, we train the reward-conditional GFlowNet with stochastic gradient $\mathbb{E}_{\tau \sim \pi_{\Theta}(\tau|s_e)}\left[\nabla_\Theta \mathcal{L}_\Theta(\tau|s_e)\right]$
% \begin{equation}
%     \mathbb{E}_{\tau \sim \pi_{\Theta}(\tau|s_e)}\left[\nabla_\Theta \mathcal{L}_\Theta(\tau|s_e)\right]
% \end{equation}
with some training trajectory distribution $\pi_{\Theta}(\tau)$. 
Akin to RL settings, we take $\pi_{\Theta}$ to be the distribution over trajectories sampled from a tempered version of current forward policy $P_{F|e}(\cdot|s_g, s_e)$.
That is, $\tau$ is sampled with $\mathbf{s}_g^{t+1} \sim P_{F|e}(\cdot|s_g^t, s_e)$ starting from $s_e^0$, mixed with a uniform action policy to ensure $\pi_\Theta$ has full support.\\

\noindent\textbf{Learning about total flow $Z$.}
Experiments show that learning the scalar function $Z(\cdot)$ end-to-end is very difficult. 
Since $Z$ represents the total flow in the entire flow network, many samples are required for an accurate estimation. Unlike the original work of trajectory balance~\citep{malkin2022trajectory}, in our setting, the scalar function $Z$ needs to condition on the external environmental state $s_e$ thus has higher sample complexity.
Interestingly, since the target EBM of GFlowNets is derived from the PRL framework in our method, $Z$ has an additional physical meaning, i.e., the soft value function $V_{\mathrm{soft}}^*(\cdot)$ in $\S$\ref{sec:prl}.
Since the soft value function is dependent on the soft $Q$ value, $Z$ can be updated by a mechanism similar to the bootstrap learning adopted by RL, thereby improving the sample efficiency.
To this end, in addition to end-to-end training of $Z$ using Equation~(\ref{eq:tbloss}), we estimate $V_{\mathrm{soft}}^*(\cdot)$ in the same way as in~\citet{sql} and fit $Z$ to it. 
The experimental results show that this form of mixed gradient update can improve the learning efficiency of Z.

\subsection{Joint Training with EBM}

Reward-conditional GFlowNets' training relies on a given function $R_{g|e}(a_e|s_g, s_e)$ to provide reward signals.
However, in the PRL framework, the energy-based policy distribution is also constantly changing with the update of the soft Q function.
Therefore, we propose a joint training framework (Algorithm~\ref{alg:dpo}), where the EBM and the reward-conditional GFlowNet are optimized alternately:
\begin{enumerate}[label=(\arabic*),leftmargin=13pt]
    \item \textbf{GFlowNet updating step: } the soft $Q$ function serves as the reward function for the GFlowNet, which is trained with the trajectory balance objective to sample from the evolving EBM;
    \item \textbf{EBM updating step: } the EBM is trained with soft $Q$ iteration~\citep[$\S$3.1]{sql} where the GFlowNet provides diverse samples.
\end{enumerate}
Moreover, again inspired by soft $Q$-learning~\citep{sql}, we find it advantageous to evaluate the forward policy, backward policy and total flow function in (\ref{eq:tbloss}) with a separate target network, where the parameters $\bar{\theta}_F$, $\bar{\theta}_B$ and $\bar{\theta}_Z$ are updated softly~\citep{lillicrap2015continuous}.

\begin{algorithm}[ht]
    \caption{Joint Training Framework of \textit{DPO}}
    \label{alg:dpo}
    \begin{algorithmic}[1]
        \STATE $\{\theta_{Q}, \theta_{F}, \theta_{B}, \theta_{Z}\}\sim$ some initialization distributions, assign target parameters $\{\bar{\theta}_{Q}, \bar{\theta}_{F}, \bar{\theta}_{B}, \bar{\theta}_{Z}\}$, $\mathcal{D} \leftarrow \text{empty replay buffer}$;
        \FOR{each epoch until some convergence conditions}
            \FOR{each timestep $t$ until the maximum limitation}
                \STATE Sample an structured action $a^t_{e}$ via $P_{F|e}(\cdot|\cdot,s^t_{e};\theta_F)$;
                \STATE Save the new experience: $\mathcal{D} \leftarrow \mathcal{D} \cup\left\{\left(s^t_{e}, a^t_{e}, r^t_{e}, s^{t+1}_{e}\right)\right\}$;
                \STATE Sample a minibatch: $\{(s_e^{(i)}, a_e^{(i)}, r_e^{(i)}, {s_e^{\prime}}^{(i)})\}_{i=0}^N \sim \mathcal{D}$.
                \STATE \textbf{EBM updating step:}
                \STATE $\quad$ Update $\theta_{Q}$ according to computed empirical gradient in (\ref{eq: sql0-policy}) and empirical soft values in (\ref{eq:soft-value});
                \STATE \textbf{GFlowNet updating step:} 
                \STATE $\quad$ Update $\{\theta_{F}, \theta_{B}, \theta_{Z}\}$ with computed empirical gradient of (\ref{eq:tbloss}), update $\theta_{Z}$ with MSE loss with computed empirical soft values additionally; 
                \STATE Update target parameters similar with~\citet{lillicrap2015continuous}.
            \ENDFOR
        \ENDFOR
    \end{algorithmic}  
\end{algorithm}

%%%%%%%%%%%%%%%%%%%%%%%%%%%%%%%%%%%%%%%%%%%%%%%%%%%%%%%%%%%%%%%%%%%%%%%%

\section{Experiments}

In this section, we will empirically validate DPO on two RL problems with structured action space, which include ATSC tasks~\citep{ault2021reinforcement} where atomic actions have \textit{physical} local dependencies;
and more generally, Battle scenarios~\citep{zheng2018magent} where atomic actions have \textit{logical} local dependencies (see Appendix for more environment details).
It is worth noting that we did not use the \textit{population diversity} (PD) proposed by~\citet{parker2020effective} or the modified PD proposed by~\citet{zhou2021continuously} as one of the evaluation metrics. 
In our experiments, we find that due to the high dimensionality and local dependencies of structured actions, PD, a locality indicator, cannot well reflect the diversity of policies.
Therefore, we evaluate different global metrics for different tasks to verify the diversity.

\subsection{Adaptive Signal Traffic Control}

We choose the following algorithms as baselines, mainly including the state-of-the-art methods for the ASTC task and for encouraging policy diversity:
\textbf{Max-Pressure} control (MP) where the phase combination with the maximal joint pressure is enabled as described in~\citep{chen2020toward};
\textbf{MPLight}-implementation is based on the FRAP open source implementation~\citep{zheng2019learning} along with the ChainerRL~\citep{fujita2021chainerrl} DQN implementation and pressure sensing;
\textbf{DvD}~\citep{parker2020effective} is a population-based RL method for effective diversity;
\textbf{SQL}~\citep{sql} method is the skeleton of the proposed DPO, which can obtain diverse policies in the low-dimensional continuous action space;
Recent proposed \textbf{RSPO}~\citep{zhou2021continuously} transforms the problem of seeking diversity policies into a constrained Markov decision process.\\

\begin{table}[htb!]
\caption{Performance ($\downarrow$) on the ATSC benchmark.}
\label{tab:astc-joint}
\resizebox{0.8\columnwidth}{!}{%
\begin{tabular}{lll|lll}
\hline
\textit{\textbf{MP}}      & Ing. Reg. & Col. Reg. & \textit{\textbf{DvD}}  & Ing. Reg.       & Col. Reg.      \\ \hline
Avg. Delay                & 59.64     & 22.06     & Avg. Delay             & 73.22           & 55.91          \\
Avg. Trip Time            & 197.23    & 86.02     & Avg. Trip Time         & 212.81          & 115.54         \\
Avg. Wait                 & 20.19     & 5.46      & Avg. Wait              & 31.36           & 28.35          \\
Avg. Queue                & 0.8       & 0.38      & Avg. Queue             & 1.42            & 2.28          \\ \hline
\textit{\textbf{SQL}}     & Ing. Reg. & Col. Reg. & \textit{\textbf{RSPO}} & Ing. Reg.       & Col. Reg.      \\ \hline
Avg. Delay                & 67.65     & 58.32     & Avg. Delay             & 90.42           & 57.28          \\
Avg. Trip Time            & 205.44    & 116.29     & Avg. Trip Time         & 226.5           & 120.53        \\
Avg. Wait                 & 26.45     & 30.01      & Avg. Wait              & 44.16           & 28.19         \\
Avg. Queue                & 1.15      & 2.06       & Avg. Queue             & 1.74            & 2.59        \\ \hline
\textit{\textbf{MPLight}} & Ing. Reg. & Col. Reg. & \textit{\textbf{DPO}}  & Ing. Reg.       & Col. Reg.      \\ \hline
Avg. Delay                & 78.16     & 60.42     & Avg. Delay             & \textbf{57.2}   & \textbf{20.28} \\
Avg. Trip Time            & 215.72    & 123.93    & Avg. Trip Time         & \textbf{192.75} & \textbf{81.42} \\
Avg. Wait                 & 34.57     & 30.34     & Avg. Wait              & \textbf{18.26}  & \textbf{4.77}  \\
Avg. Queue                & 1.48      & 2.33      & Avg. Queue             & \textbf{0.65}   & \textbf{0.32}  \\ \hline
\end{tabular}%
}

% \vspace{-15pt}
\end{table}

\begin{figure}[htb!]
    \centering
    \begin{subfigure}[b]{0.23\textwidth}
        \centering
        \includegraphics[width=\textwidth]{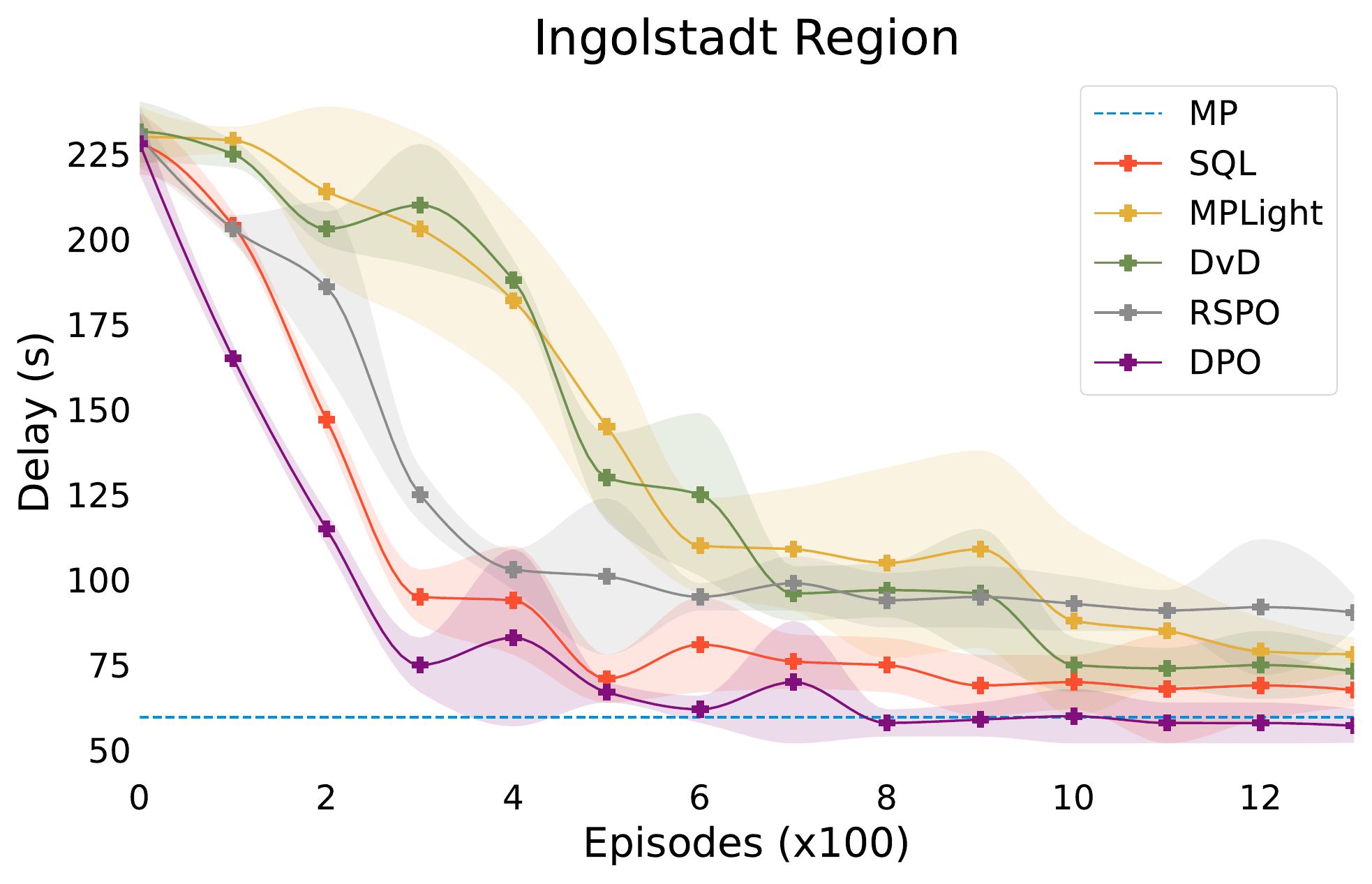}
    \end{subfigure}
    \begin{subfigure}[b]{0.23\textwidth}
        \centering
        \includegraphics[width=\textwidth]{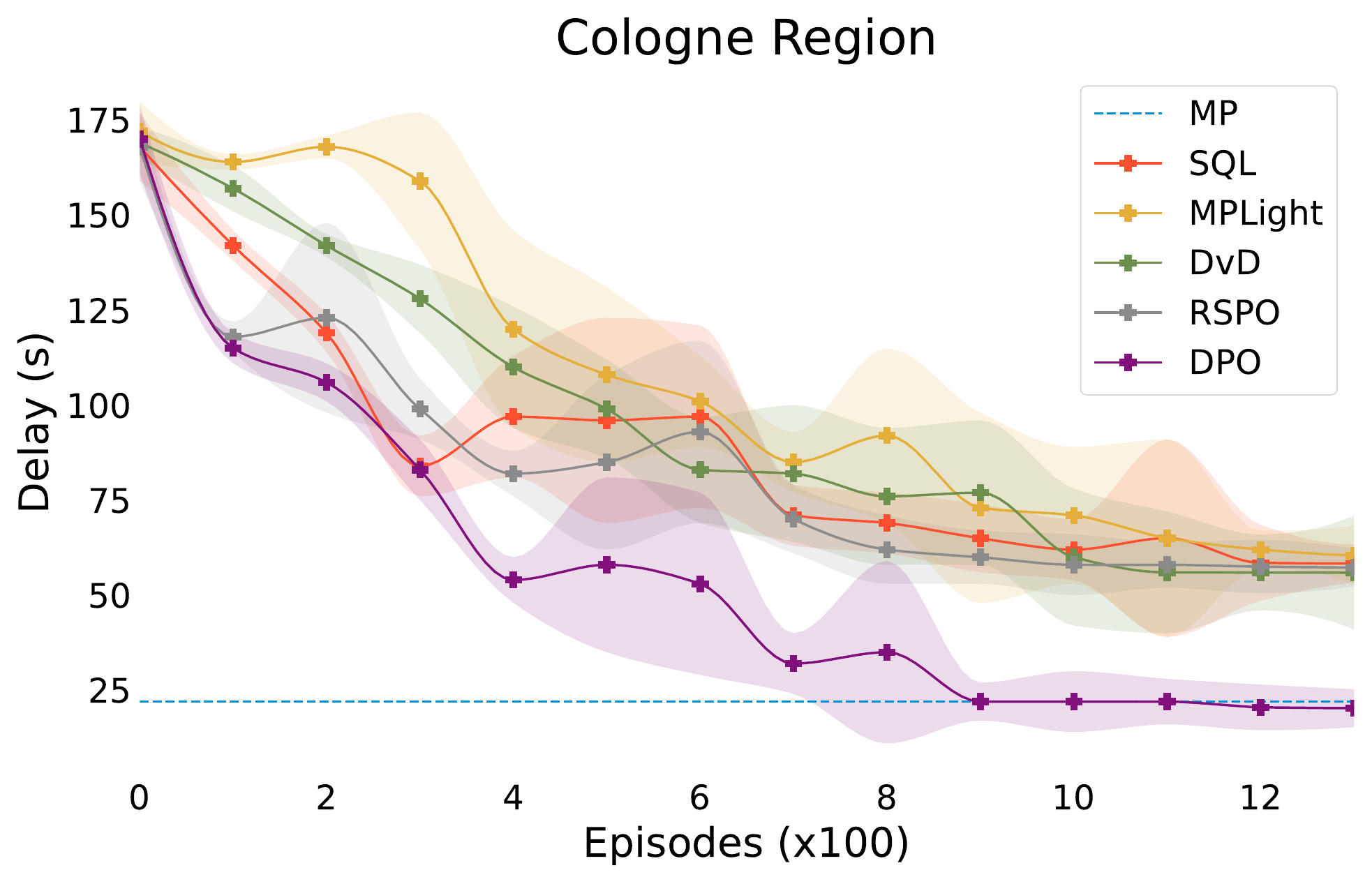}
    \end{subfigure}
    \caption{Learning curves of decay ($\downarrow$) on the ATSC.}
    \label{fig:atsc-decay}
    \vspace{-10pt}
\end{figure}

From the experimental results in Table~\ref{tab:astc-joint} and Figure~\ref{fig:atsc-decay}, it can be seen that DPO achieves state-of-the-art (SOTA) performance and convergence speed on two coordinated control tasks in TAPAS Cologne and InTAS scenarios.
It is worth noting that classical MP methods based on heuristic rules and expert knowledge also show good results.
DPO can outperform the MP method through a reinforcement learning mechanism, showing its superiority in solving the ATSC problem.
While among the three algorithms that encourage policy diversity, the DvD performs the worst, which we believe is due to the limitations of how it computes the distance between two policies on complex problems.
The other two algorithms, SQL and RSPO, can show near-SOTA performance on small-scale problems, i.e., the TAPAS Cologne scenario where a structured action consists of $8$ atomic actions. 
However, in the larger-scale InTAS scenario, its performance drops sharply, which shows that existing algorithms that encourage policy diversity have certain limitations when dealing with structured action spaces.

\begin{figure}[htb!]
    \centering
    \includegraphics[width=0.45\textwidth]{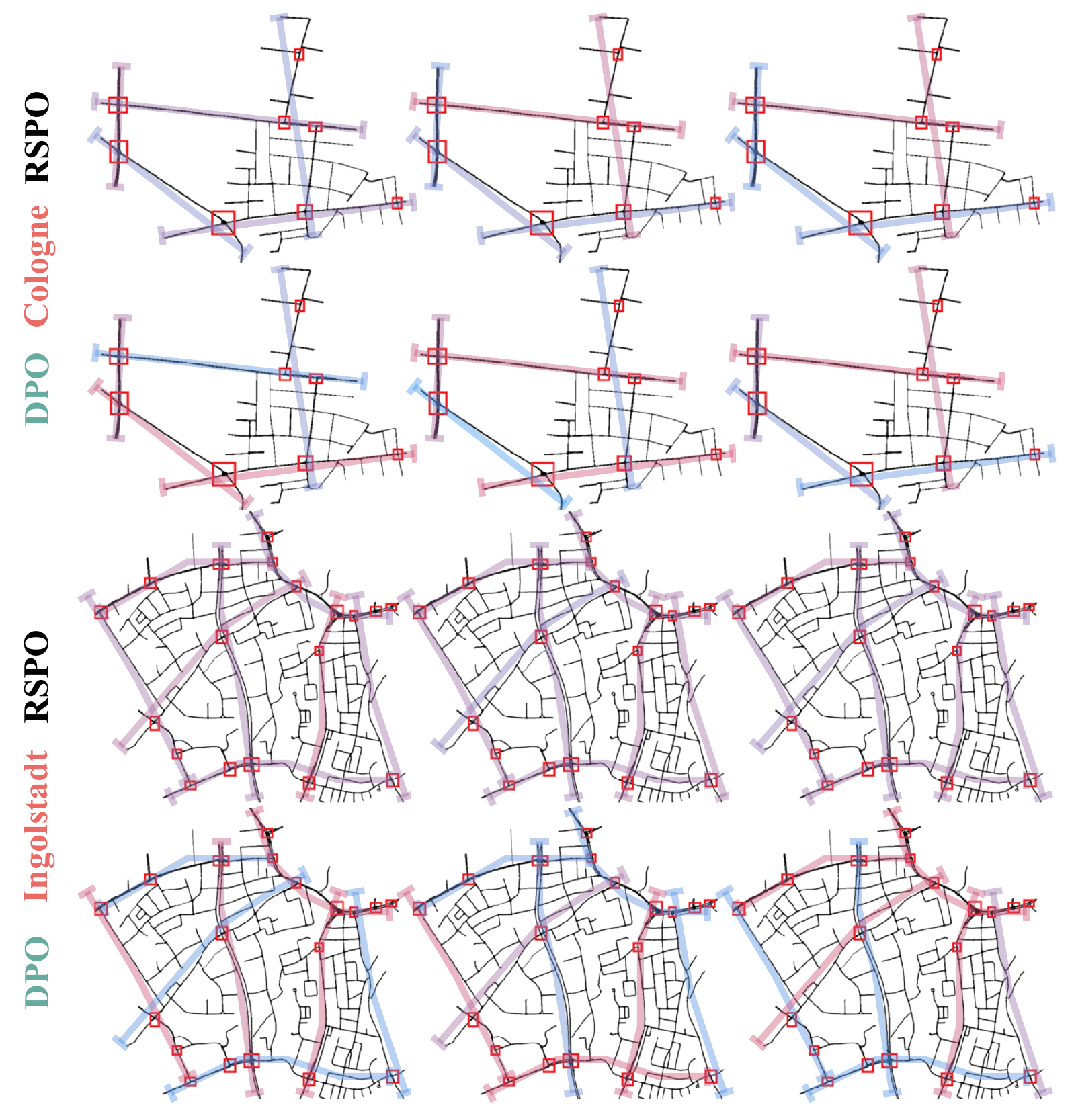}
    \caption{Comparison of policy diversity between DPO and RSPO under the ATSC benchmark. Different colors represent different commute times.}
    \label{fig:atsc-div-main}
    \vspace{-10pt}
\end{figure}

Figure~\ref{fig:atsc-div-main} shows the comparison of the policy diversity between RSPO and DPO (see the appendix for more results).
We ignore the atomic action level, that is, the diversity of each traffic light's phase selection strategy, but the diversity of the entire road network's traffic control strategy. 
To this end, we calculate the average commute time of the main road under multiple random seeds for different algorithms in different scenarios. 
Furthermore, for visualization convenience, we normalized each algorithm separately. Red indicates longer commute time; otherwise, it is shown in blue.
As seen from the figure, DPO learns policies with sufficient diversity in structured action spaces of different scales, but RSPO only shows some effect in small-scale tasks.

\subsection{Battle Scenario}

In the Battle scenario, the atomic action is each agent's action, and we transform the logical correlation between each agent into the physical correlation without loss of generality.
Expressly, we assume that atomic actions with local logical correlations have a fixed influence range with a radius $d=4$ in Euclidean space.
% An atomic action can then establish a physical correlation with others within its influence range.
In this benchmark, we additionally select \textbf{IDQN}, the built-in algorithm in the MAgent, and \textbf{MFQ}~\citep{yang2018mean}, the state-of-the-art algorithm on the Battle as baselines.

\begin{figure}[htb!]
    \centering
    \begin{subfigure}[b]{0.23\textwidth}
        \centering
        \includegraphics[width=\textwidth]{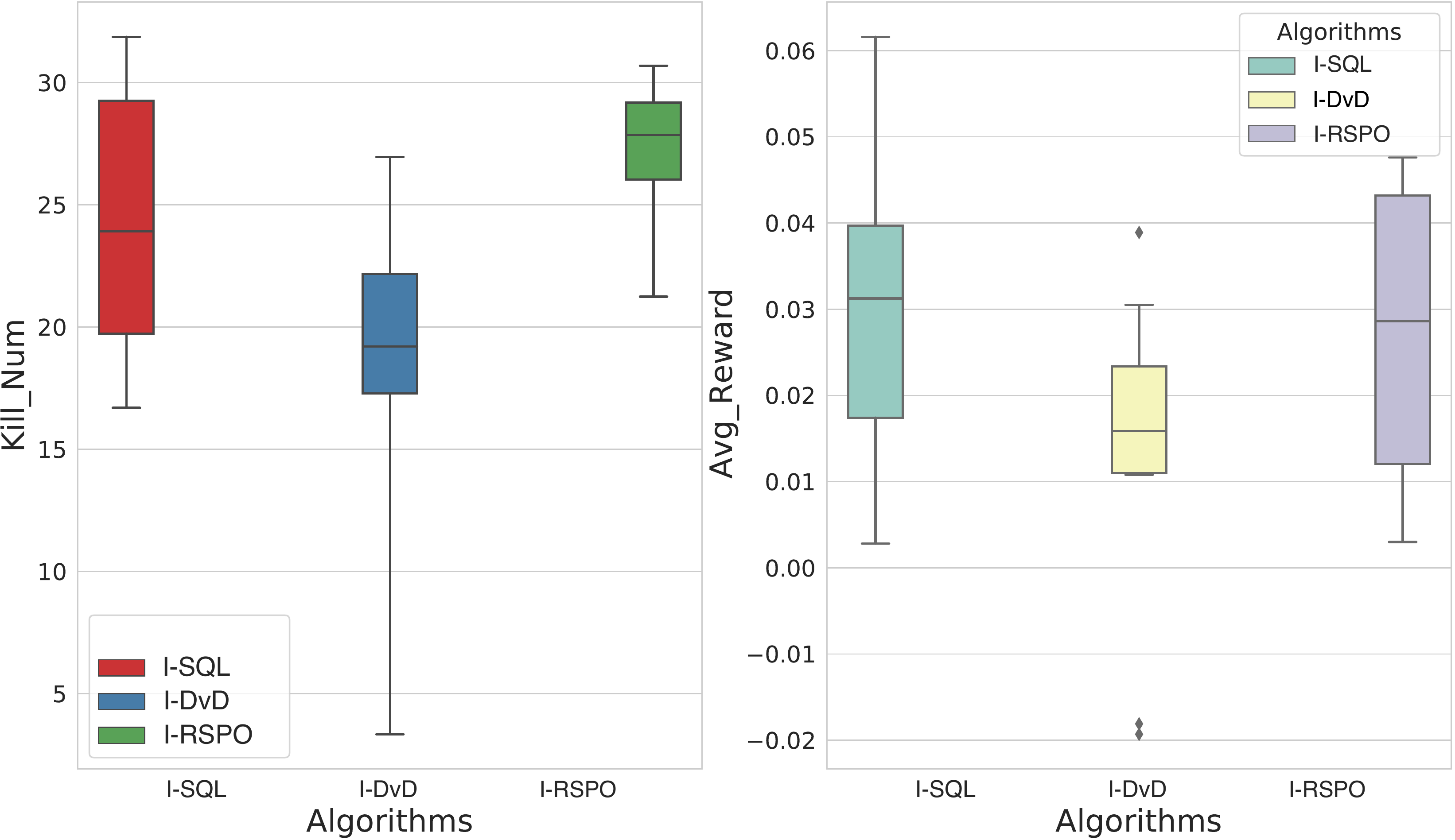}
        % \caption{}
    \end{subfigure}
    \begin{subfigure}[b]{0.23\textwidth}
        \centering
        \includegraphics[width=\textwidth]{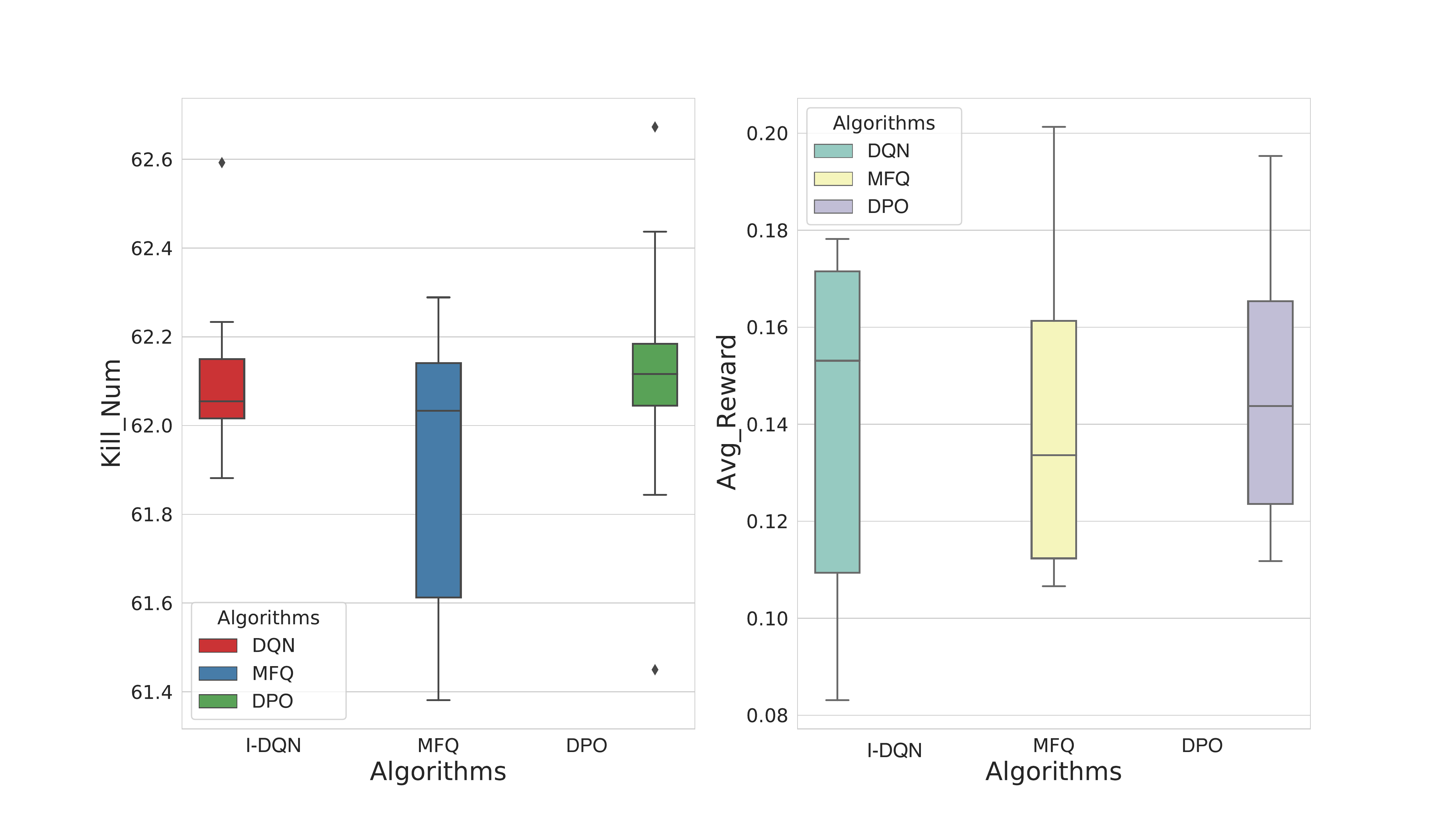}
        % \caption{}
    \end{subfigure}
    \caption{Boxplot of average kill number ($\uparrow$) and average agent reward ($\uparrow$) of $50$ runs on Battle Game. Results compares the average agent number of blue army killed by red army (left part of each figure) and the average individual rewards of each agent (right part of each figure) respectively.}
    \label{fig:battle-joint}
    % \vspace{-5pt}
\end{figure}

We first train the IDQN in a self-play way
% and save its converged checkpoints.
and the blue agent loads the checkpoint 
% of the above IDQN algorithm 
and fixes the model parameters. 
The red agent is then trained with different algorithms, and the final result is shown in Figure~\ref{fig:battle-joint}.
It is worth noting that DvD, SQL, and RSPO are less scalable.
% when dealing with sizeable structured action spaces.
So in the Battle scenario, we combine independent learning to obtain I-DvD, I-SQL, and I-RSPO variants.
Independent learning does not constrain the algorithm's performance, while the IDQN algorithm also shows promising results.
As seen from the figure, the three algorithms that encourage policy diversity do not show good results in large-scale structured action spaces, while DPO can still stably approach the performance of SOTA.

\begin{figure}[htb!]
    \centering
    \includegraphics[width=0.4\textwidth]{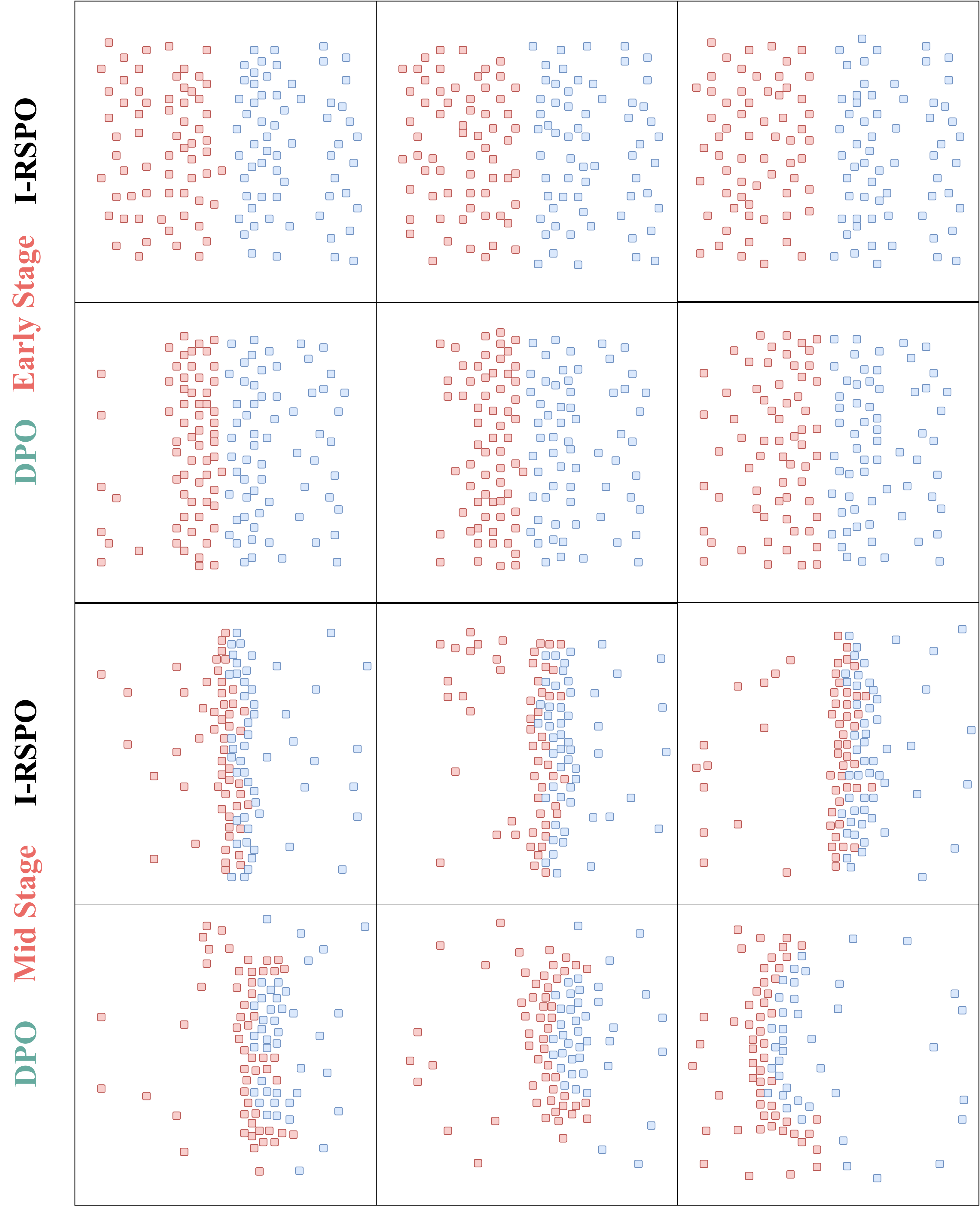}
    \caption{Comparison of policy diversity between DPO and RSPO under the early and middle stages of the Battle.}
    \label{fig:battle-div-main}
    \vspace{5pt}
\end{figure}

Figure~\ref{fig:battle-div-main} shows the diversity of policies between I-RSPO and DPO in the early and middle stages of the game (see appendix for more results). 
As seen from the figure, the policies learned by DPO show a variety of deployment strategies in the early stage; in the middle stage, the enemy can be surrounded by different formations to maximize the attack power. 
Although I-RSPO based on independent learning shows a specific diversity at the individual level, it is not easy to generate different policies as a whole.

\begin{figure}[htb!]
    \centering
    \begin{subfigure}[c]{0.23\textwidth}
        \centering
        \includegraphics[width=\textwidth]{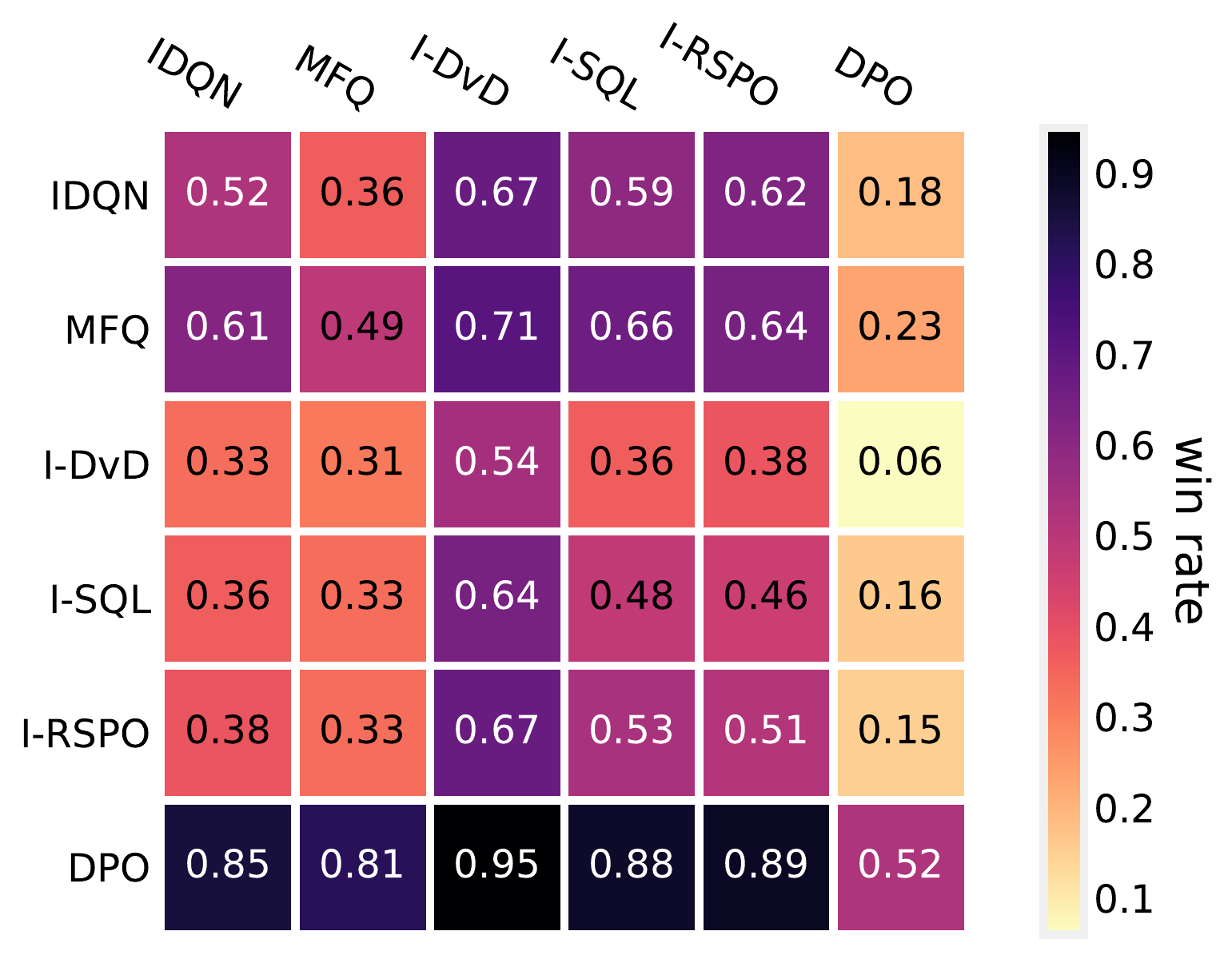}
        % \caption{}
    \end{subfigure}
    \begin{subfigure}[c]{0.23\textwidth}
        \centering
        \includegraphics[width=\textwidth]{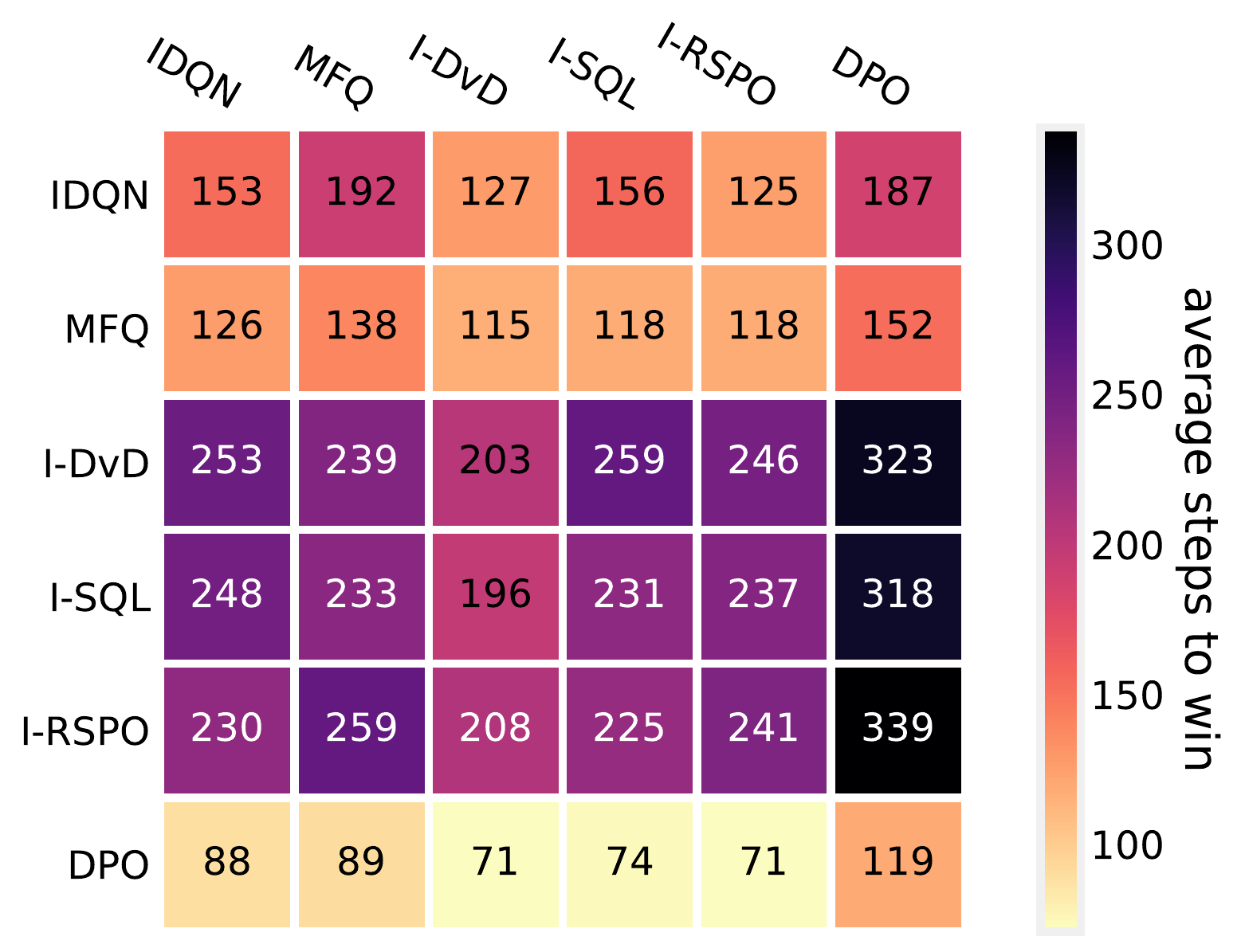}
        % \caption{}
    \end{subfigure}
    \caption{The heatmap of the (Left) win ratio ($\uparrow$) and (Right) average steps to win ($\downarrow$) among DPO and others of the testing phase of the Battle benchmark.}
    % (Right) Partial view of the battle game between DPO (red) and MDQ (blue).}
    \label{fig:battle-heatmap}
\end{figure}

Diverse policies are more difficult to be exploited by opponents in competitive scenarios and can better adapt to changes in opponents' policies. 
In order to verify the above point, we let the red agents trained based on different algorithms compete against each other and count the average winning rate. 
The results are shown in Figure~\ref{fig:battle-heatmap}.
As seen from the figure, DPO shows good robustness against different opponents.

\section{Closing Remarks}

In this paper, we aim to seek diverse policies in an under-explored setting, namely RL tasks with \textit{structured action spaces} with the \textit{composability} and \textit{local dependencies}.
The complex action structure, non-uniform reward landscape, and subtle hyperparameter tuning due to the structured actions prevent existing methods from scaling well.
We propose a simple and effective method, \textit{Diverse Policy Optimization (DPO)}, to model the policies in structured action space as the energy-based models by following the probabilistic RL framework.
DPO adopts a joint training framework, where the energy-based model, and the generative flow  network, which is introduced as the efficient, diverse EBM-based policy sampler, are optimized alternately:
The energy function serves as the negative log-reward function for the GFlowNet, which is trained with the trajectory balance objective to sample from the evolving energy-based policies. 
In contrast, the energy function is trained with soft Bellman backup, where the GFlowNet provides diverse samples.
Experiments demonstrate that the proposed DPO is both general and practical across structured action spaces with physical and, more generally, logical local dependencies.

%%%%%%%%%%%%%%%%%%%%%%%%%%%%%%%%%%%%%%%%%%%%%%%%%%%%%%%%%%%%%%%%%%%%%%%%

%%% The acknowledgments section is defined using the "acks" environment
%%% (rather than an unnumbered section). The use of this environment 
%%% ensures the proper identification of the section in the article 
%%% metadata as well as the consistent spelling of the heading.

\begin{acks}
This work was supported in part by Postdoctoral Science Foundation of China (2022M723039), NSFC (62106213, 72150002), SSTP (RCBS20210609104356063, JCYJ20210324120011032), and a grant from Shenzhen Institute of Artificial Intelligence and Robotics for Society.
% If you wish to include any acknowledgments in your paper (e.g., to 
% people or funding agencies), please do so using the `\texttt{acks}' 
% environment. Note that the text of your acknowledgments will be omitted
% if you compile your document with the `\texttt{anonymous}' option.
\end{acks}

%%%%%%%%%%%%%%%%%%%%%%%%%%%%%%%%%%%%%%%%%%%%%%%%%%%%%%%%%%%%%%%%%%%%%%%%

%%% The next two lines define, first, the bibliography style to be 
%%% applied, and, second, the bibliography file to be used.

\balance

\bibliographystyle{ims} 
\bibliography{sample}

\clearpage
\newpage

\appendix

\onecolumn

\centerline{\Huge \textbf{Supplementary Material for}}\hfill \break
\centerline{\Huge \textbf{Diverse Policy Optimization for Structured Action Space}}\hfill \break

\section{Related Works}

To the best of our knowledge, existing work on reinforcement learning rarely pursues both the quality as well as the diversity of optimal policies in sequential decision problems with large-scale, structured action spaces.
Therefore, this section will briefly review the work in reinforcement learning focusing on the diversity of solutions and dealing with sequential decision problems with large-scale or structured action spaces, respectively.

\subsection{Diverse Solutions in RL}

Most of the literature on this problem has been done in the field of neuroevolution methods inspired by Quality-Diversity (QD), seeking to maximize the reward of a policy through approaches strongly motivated by natural biological processes. 
They typically work by perturbing a policy and either computing a gradient (as in Evolution Strategies) or selecting the top-performing perturbations (as in Genetic Algorithms). 
Neuroevolution methods comprise two leading families of algorithms: MAP-Elites~\citep{cully2015robots,mouret2015illuminating} and novelty search with local competition~\citep{Lehman2011EvolvingAD}. 
These methods typically maintain a collection of policies and adapt it using evolutionary algorithms to balance the QD trade-off~\citep{pugh2016quality,duarte2017evolution,parker2020effective,nilsson2021policy,gangwani2021harnessing,Lim2022DynamicsAwareQF}.

In another part of the work, intrinsic rewards have been used for learning diversity in terms of the discriminability of different trajectory-specific quantities~\citep{gregor2016variational,eysenbach2018diversity,hartikainen2019dynamical,goyal2019reinforcement,sharma2020emergent,zahavy2020planning,alver2021constructing}.
These methods are similar in principle to novelty search without a reward signal but instead focus on diversity in behaviors defined by the states they visit.
Other work implicitly induces diversity to learn policies that maximize the set robustness to the worst-possible reward~\citep{kumar2020one,Zahavy2021DiscoveringAS}, or uses diversity as a regularizer when maximizing the extrinsic reward~\citep{Levine2018ReinforcementLA,gangwani2018learning,masood2019diversity,sharma2019dynamics,zhang2019learning}.
There is also a small body of work that transforms the problem of seeking diversity policies into a Constrained Markov Decision Process~\citep{sun2020novel,zhou2021continuously,Derek2021AdaptableAP,Zahavy2022DiscoveringPW}.

In addition to getting policies with diversity in RL, some related work is encouraging policy diversity.
In imitation learning, the problem of imitating diverse behaviors from expert demonstrations has been addressed in previous studies~\citep{wang2017robust,li2017infogail,sharma2018directed,merel2018neural}. 
In these methods, diverse behaviors are encoded in latent variables. 
However, these imitation learning methods assume the availability of observations of diverse behaviors performed by experts.
Encouraging agents to diversify their exploration in the early stages of RL has also received significant attention in recent years~\citep{hong2018diversity,conti2018improving,khadka2019collaborative,liu2018emergent,majumdar2020evolutionary,peng2020non}.
The diversity of policies in multi-agent reinforcement learning (MARL) is also crucial to improve the agent's robustness and their ability to zero-shot cooperate~\citep{tang2020discovering,yang2020multi,Lupu2021TrajectoryDF,Nieves2021ModellingBD}.

% \begin{itemize}
%     % \item QD algorithm~\citep{pugh2016quality,duarte2017evolution,parker2020effective,nilsson2021policy,gangwani2021harnessing,Lim2022DynamicsAwareQF}
%     %     \begin{itemize}
%     %         \item MAP-Lite~\citep{cully2015robots,mouret2015illuminating}
%     %         \item local competition~\citep{Lehman2011EvolvingAD}
%     %     \end{itemize}
%     \item exploration~\citep{hong2018diversity,conti2018improving,khadka2019collaborative,liu2018emergent,majumdar2020evolutionary,peng2020non}
%     % \item intrinsic reward
%     %     \begin{itemize}
%     %         \item discriminability~\citep{gregor2016variational,eysenbach2018diversity,hartikainen2019dynamical,goyal2019reinforcement,sharma2020emergent,zahavy2020planning,alver2021constructing}
%     %         \item regulizer~\citep{Levine2018ReinforcementLA,gangwani2018learning,masood2019diversity,sharma2019dynamics,zhang2019learning}
%     %     \end{itemize}
%     % \item learning from expert~\citep{wang2017robust,li2017infogail,sharma2018directed,merel2018neural}
%     % \item constraint MDP~\citep{sun2020novel,zhou2021continuously,Derek2021AdaptableAP,Zahavy2022DiscoveringPW}
%     % \item worst-case~\citep{kumar2020one,Zahavy2021DiscoveringAS}
%     \item multi-agent reinforcement learning~\citep{tang2020discovering,yang2020multi,Lupu2021TrajectoryDF,Nieves2021ModellingBD}
% \end{itemize}

\subsection{Structured or Large-Scale Actions}

A large part of the current work on policy optimization for structured action spaces addresses one particular class of problems, namely, parametric action space problems, in which the action space has a particular master-slave structure. 
% The master action space is a discrete action space in which each action corresponds to a continuous action in a continuous slave action space. 
% For example, the autonomous driving task is a typical task with a parametric action space. 
% Assuming that the direction in which the car travels is discrete. The direction corresponds to the discrete master action space, and the duration of travel in that direction is the continuous slave action space, which is a continuous parameter of the discrete direction action.
The difficulty in solving the parameterized action space lies in the heterogeneity of discrete master actions and continuous slave actions. 
Current methods either learn a continuous parameter policy for each discrete action~\citep{masson2016reinforcement,xiong2018parametrized,bester2019multi}; 
or discrete actions are output in parallel with continuous actions and employ gradient post-processing techniques or improved value function networks to solve the master-slave action correspondence problem~\citep{hausknecht2015deep,fan2019hybrid}; 
or first, generate discrete actions, then generate continuous parameters based on that action and design sophisticated gradient update schemes for end-to-end training~\citep{delalleau2019discrete,berner2019dota,wei2018hierarchical}.

In contrast, there are fewer algorithms oriented towards structured action spaces in general, and in the tasks solved by these algorithms, there are no explicit dependencies between atomic actions.
Thus, existing approaches are either based on the assumption of independence of the decomposed sub-actions~\citep{tavakoli2018action,mahajan2021reinforcement}; 
or they are based on the inductive bias to assign a conditional dependency structure to the decomposed sub-actions and pick up the actions one by one through an autoregressive form based on recurrent neural networks, which are finally spliced into the original actions~\citep{metz2017discrete,pierrot2020factored}. 
There are also a series of approaches that assume a game relationship between the decomposed sub-actions, model each sub-action as an agent, and use MARL methods to solve them~\citep{yang2018mean,fu2019deep,vinyals2019grandmaster,mahajan2021tesseract,li2021structured}. 
However, the field of MARL is still in the preliminary exploration stage, and numerous theoretical problems remain unsolved. 
Thus modeling as a multi-agent problem will introduce more new challenges.

% In the following, we review and summarize the existing policy optimization methods for large-scale action spaces. 
% Large-scale here means that the action space is ample in size, but the action space in these high-dimensional tasks is not structured.
To address the \textit{curse of dimensionality} caused by (non-structured) large-scale action spaces, existing methods are based on the idea of reshaping the action space and thus reducing the dimensionality, e.g., some works perform dimensionality reduction by clustering the actions~\citep{dulac2015deep,chandak2019learning,he2015deep,wang2016exploring,tennenholtz2019natural}. 
However, these approaches require the assumption that actions have dense semantic information, consist of natural language, and cannot be applied to general high-dimensional tasks. 
Some works propose solutions for generic large-scale action spaces, such as dividing the action space by using multiple hierarchical policies similar to a tree structure to reduce the action dimension of each layer of the policy~\citep{zahavy2018learn,chen2019large,delarue2020reinforcement};
or gradually increasing the action space employing curriculum learning so that the policy only needs to be optimized in a smaller action space in the early stage~\citep{farquhar2020growing}.

% %%%%%%%%%%%%%%%%%%%%%%%%%%%%%%%%%%%%%%%%%%%%%%%%%%%%%%%%%%%%%%%%%%%%%%%%

% \clearpage
% \newpage

% \section{Theoretical Results}

%%%%%%%%%%%%%%%%%%%%%%%%%%%%%%%%%%%%%%%%%%%%%%%%%%%%%%%%%%%%%%%%%%%%%%%%

\clearpage
\newpage

\section{Training Details}

\subsection{Environments}

\begin{figure}[htb!]
    \centering
    \includegraphics[width=0.48\textwidth]{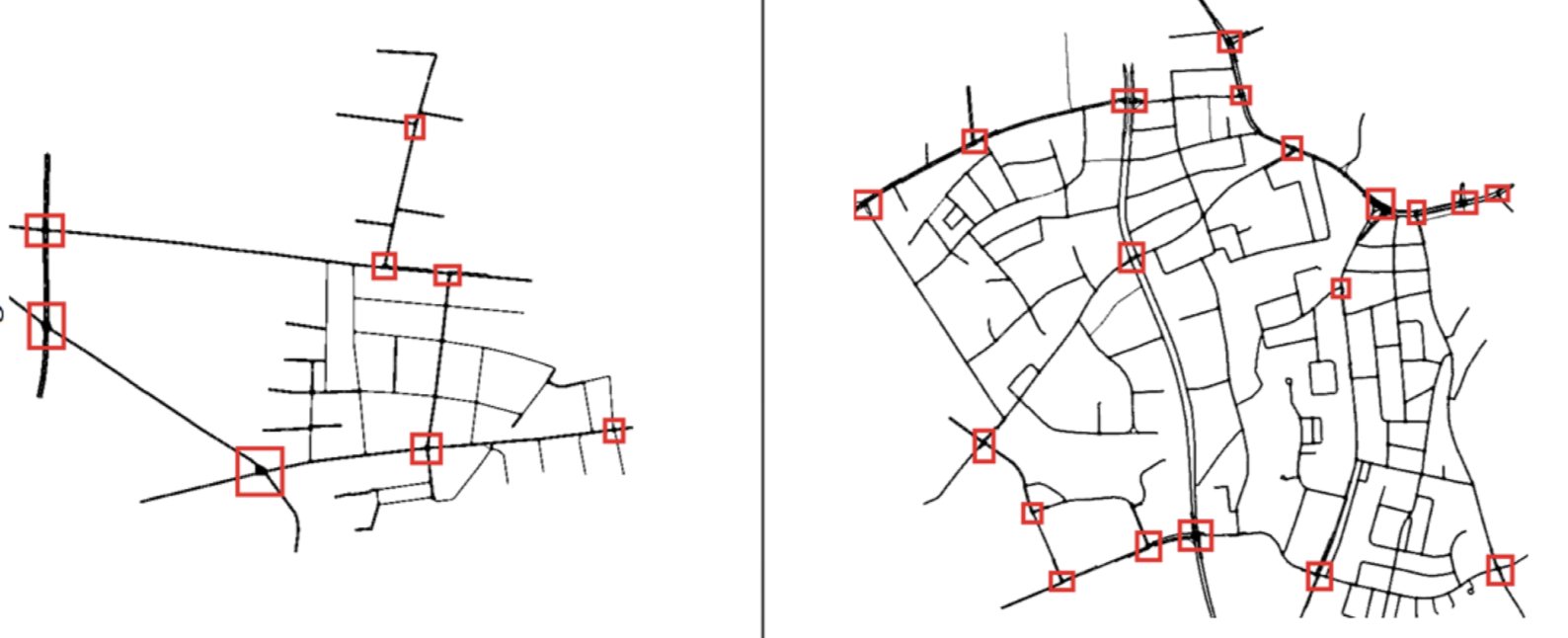}
    \caption{The road networks of coordinated control task~\citep{ault2021reinforcement} for TAPAS Cologne~\citep{varschen2006mikroskopische} and InTAS~\citep{lobo2020intas} respectively.}
    \label{fig:roadnet}
\end{figure}

\noindent\textbf{Adaptive Signal Traffic Control.} 
This benchmark based on $2$ well-established Simulation of Urban Mobility traffic simulator (SUMO)~\citep{sumo} scenarios, namely, ``TAPAS Cologne''~($8$ lights)~\citep{varschen2006mikroskopische} and ``InTAS''~($21$ lights)~\citep{lobo2020intas}, which describe traffic within a real-world city, Cologne and Ingolstadt (Germany) respectively.
There are $3$ kinds of tasks in the original work~\citep{ault2021reinforcement}, namely (a) controlling a single intersection, (b) controlling multiple intersections along an arterial corridor, and (c) coordinated control of multiple intersections within a congested area.
We select the most complex coordinated control task (c) to demonstrate the advantage of DPO in finding diverse policies.
In the coordinated control task, the atomic action is the selection of the signal light's phase at each intersection, and the physical dependencies are the roads between the intersections. 
The road network is shown in Figure~\ref{fig:roadnet}.

\begin{figure}[htb!]
    \centering
    \includegraphics[width=0.48\textwidth]{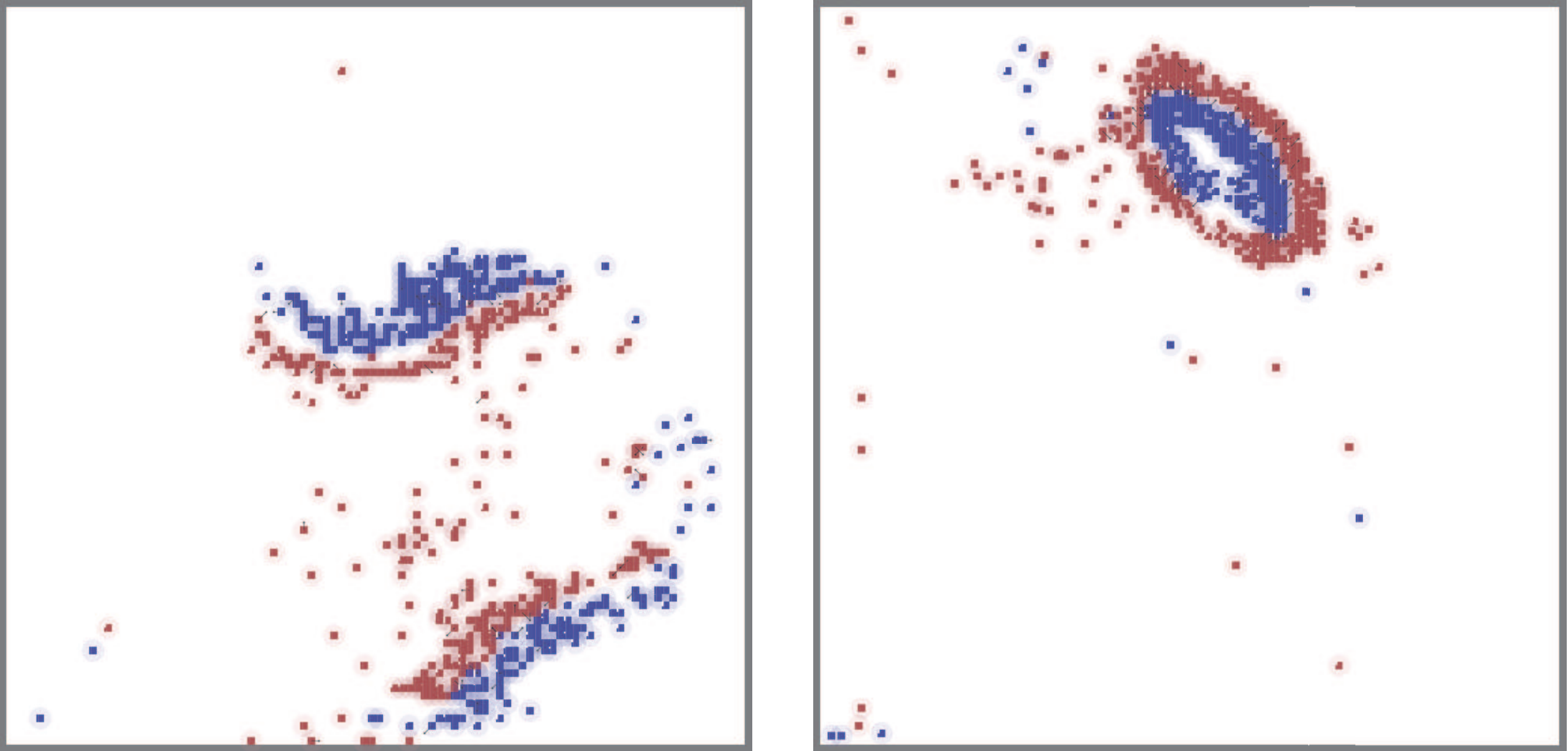}
    \caption{The snapshot of Battle in MAgent platform~\citep{zheng2018magent}.}
    \label{fig:battle}
\end{figure}

\noindent\textbf{Battle Scenario.}
This benchmark is based on the MAgent~\citep{zheng2018magent}, a research platform for many-agent reinforcement learning.
We selected the competitive task, Battle, as the simulation environment to highlight the advantages of the diverse policies. 
In Battle, $n$ agents learn to fight against $n$ enemies who have superior abilities than agents. 
(Figure~\ref{fig:battle}). 
As the enemy's hit point is more than a single agent's damage, agents must continuously cooperate to kill the enemy.

In our experiments on the ATSC and Battle benchmarks, all the environment settings, such as the definition of state, the definition of reward, etc., as well as the evaluation metrics, are kept the same as in~\citet{ault2021reinforcement}\footnote{\url{https://github.com/Pi-Star-Lab/RESCO} .} and~\citet{terry2021pettingzoo}\footnote{\url{https://github.com/Farama-Foundation/PettingZoo}.} respectively.

\subsection{Methods}

\noindent\textbf{Random seeds.} 
Except as mentioned in the text, all experiments were run for $5$ random seeds each. 
Graphs show the average (solid line) and std dev (shaded) performance over random seed throughout training.
In the ATSC benchmark, the tables show the empirical mean of the relevant evaluation metrics.\\

\noindent\textbf{Hyperparameters.}
Table~\ref{table:default} shows the tuning range of hyperparameters used for all the experiments of our method and baselines.
For all hyperparameters that need to be tuned, we use the \textit{Bayesian hyperparameter search} method in the wandb platform\footnote{\url{https://wandb.ai/}} for parallel tuning.
During the parallel tuning, the platform will create a probabilistic model of a metric score as a function of the hyperparameters, and choose parameters with high probability of improving the metric. 
Bayesian hyperparameter search method uses a Gaussian Process to model the relationship between the parameters and the model metric and chooses parameters to optimize the probability of improvement.
\\

\noindent\textbf{Hardware.} The hardwares used in the experiment are a server with $128$ cores, $128$G memory and $4$ NVIDIA GeForce RTX 1080Ti graphics cards with $11$G video memory, and a server with $128$ cores, $256$G memory and $2$ NVIDIA GeForce RTX 3090 graphics cards with $24$G video memory.\\

\noindent\textbf{The Code of Baselines.}
The code and license of baselines are shown in following list:
\begin{itemize}
    \item IDQN~\citep{zheng2018magent}: \url{https://github.com/geek-ai/MAgent}, MIT License;
    \item MFQ~\citep{yang2018mean}: \url{https://github.com/mlii/mfrl}, MIT License;
    \item Max-Pressure~\citep{chen2020toward}: \url{https://github.com/Pi-Star-Lab/RESCO}, No License;
    \item MPLight~\citep{zheng2019learning}: \url{https://github.com/Pi-Star-Lab/RESCO}, No License;
    \item DvD~\citep{parker2020effective}: \url{https://github.com/jparkerholder/DvD_ES}, Apache-2.0 license;
    \item SQL~\citep{sql}: \url{https://github.com/haarnoja/softqlearning}, No License;
    \item RSPO~\citep{zhou2021continuously}: \url{https://github.com/footoredo/rspo-iclr-2022}, No License.
\end{itemize}

\begin{table*}[h]
  \caption{Hyperparameters of all methods used in experiments.}
  \label{table:default}
  \begin{subtable}[t]{0.48\textwidth}
  \centering
  \resizebox{\textwidth}{!}{%
  \begin{tabular}{lr}
    \toprule
    Name & Tuning Range \\
    \midrule
    \texttt{number of GPN layers} & 3\\
    \texttt{hidden units of GPN} & \{64, 128, 256\}\\
    \texttt{dropout of GPN layers} & 0.6\\
    \texttt{dropout of GPN attention layer} & 0.5\\
    \texttt{alpha of GPN} & (0, 1)\\
    \texttt{number of heads of GPN attention layer} & 4\\
    \texttt{use residual in GPN} & \texttt{True}\\
    \texttt{norm layer in GPN} & \{\texttt{Layernorm, Batchnorm}\}\\
    \texttt{number of GAT layers (soft-Q)} & 3\\
    \texttt{hidden units of GAT (soft-Q)} & \{64, 128, 256\}\\
    \texttt{dropout of GAT layers (soft-Q)} & 0.6\\
    \texttt{dropout of GAT attention layer (soft-Q)} & 0.5\\
    \texttt{alpha of GAT (soft-Q)} & (0, 1)\\
    \texttt{number of heads of GAT attention layer (soft-Q)} & 4\\
    \texttt{use residual in GAT (soft-Q)} & \texttt{True}\\
    \texttt{norm layer in GAT (soft-Q)} & \{\texttt{Layernorm, Batchnorm}\}\\
    \texttt{learning rate of GPN} & (1e-5, 1e-3) \\
    \texttt{learning rate of $Z$} & (1e-3, 1e-1) \\
    \texttt{learning rate of GAT} & (1e-5, 1e-3) \\
    \texttt{Optimizers} & \texttt{AdamW} \\
    \texttt{Replay Buffer Size} & 1e6 \\
    \texttt{$\gamma$} & (0.9, 0.99) \\
    \texttt{replay start size} & 32 \\
    \texttt{minibatch size} & 32 \\
    \texttt{max gradient norm} & 20 \\
    \texttt{initial temperature (soft-Q)} & 1.0 \\
    \texttt{temperature learning rate} & (1e-5, 3e-4) \\
    \texttt{soft update coefficient} & (2e-3, 5e-1) \\
    \texttt{GPN update ratio} & (2, 6) \\
    \texttt{number of GPN updates} & (1, 10) \\
    \bottomrule
  \end{tabular}%
  }
  \caption{DPO.}
  \end{subtable}
  \begin{subtable}[t]{0.3\textwidth}
  \centering
  \resizebox{\textwidth}{!}{%
  \begin{tabular}{lr}
    \toprule
    Name & Tuning Range \\
    \midrule
    \texttt{learning rate (IDQN)} & 1e-3\\
    \texttt{training frequency (IDQN)} & 5\\
    \texttt{batch size (IDQN)} & 256\\
    \texttt{target update (IDQN)} & 1200\\
    \texttt{memory size (IDQN)} & 2e20\\
    \texttt{learning rate (IDQN)} & 1e-4\\
    \texttt{$\gamma$ (IDQN)} & 0.99\\
    \texttt{learning rate (MFQ)} & 1e-4\\
    \texttt{exploration decay (MFQ)} & $1.0 \rightarrow 0.05, 2000$\\
    \texttt{$\gamma$ (MFQ)} & 0.95\\
    \texttt{batch size (MFQ)} & 128\\
    \texttt{memory size (MFQ)} & 5e5\\
    \texttt{batch size (MPLight)} & 32\\
    \texttt{$\gamma$ (MPLight)} & 0.99\\
    \texttt{exploration decay (MPLight)} & $1.0 \rightarrow 0.0, 220$\\
    \texttt{target update (MPLight)} & 500\\
    \texttt{demand shape (MPLight)} & 1\\
    \texttt{$\sigma$ (DvD)} & (1e-4, 1e-2)\\
    \texttt{$\eta$ (DvD)} & (1e-4, 1e-2)\\
    \texttt{hidden units (DvD)} & \{32, 64, 128\}\\
    \texttt{ES-sensings (DvD)} & \{200, 300, 400\} \\
    \texttt{$K$ (SQL)} & (32, 100)\\
    \texttt{$M$ (SQL)} & (32, 100)\\
    \texttt{$K_V$ (SQL)} & 50\\
    \texttt{$alpha$ (RSPO)} & (0.1, 1.5)\\
    \texttt{$\lambda_{\mathrm{B}}^{int}$ (RSPO)} & (0, 10)\\
    \texttt{$\lambda_{\mathrm{R}}^{int}$ (RSPO)} & (0, 1)\\
    \texttt{Initial learning rate (RSPO)} & (1e-4, 1e-3)\\
    \texttt{Batch size (RSPO)} & \{512, 1600, 6400\}\\
    \texttt{PPO epochs (RSPO)} & (1, 10)\\
    \bottomrule
  \end{tabular}%
  }
  \caption{Baselines.}
  \end{subtable}

\end{table*}

Learning curves are smoothed by the exponential moving average technique with coefficient $0.6$.
Source code is available at this anonymous code repository\footnote{\url{https://anonymous.4open.science/r/DPO}.}, which is based on~\citep{gpn}\footnote{\url{https://github.com/qiang-ma/graph-pointer-network}.} and~\citet{zhang2022generative}\footnote{\url{https://github.com/GFNOrg/EB_GFN}.}.

%%%%%%%%%%%%%%%%%%%%%%%%%%%%%%%%%%%%%%%%%%%%%%%%%%%%%%%%%%%%%%%%%%%%%%%%
\clearpage
\newpage

\section{More Results}

\begin{figure}[htb!]
    \centering
    \begin{subfigure}[b]{0.53\textwidth}
        \centering
        \includegraphics[width=\textwidth]{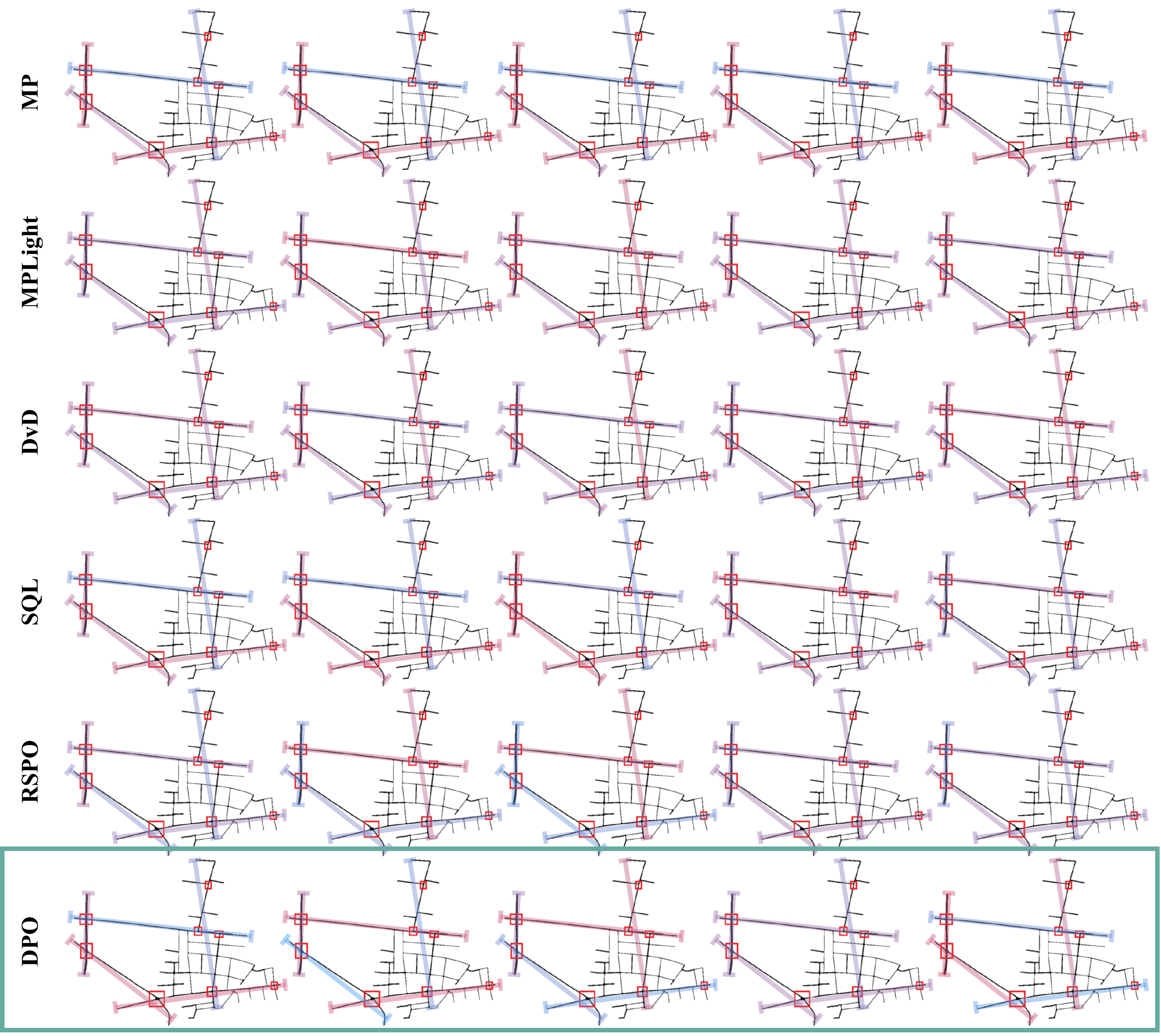}
        \caption{The Ingostadt scenario.}
        \label{fig:atsc-div}
    \end{subfigure}
    \begin{subfigure}[b]{0.45\textwidth}
        \centering
        \includegraphics[width=\textwidth]{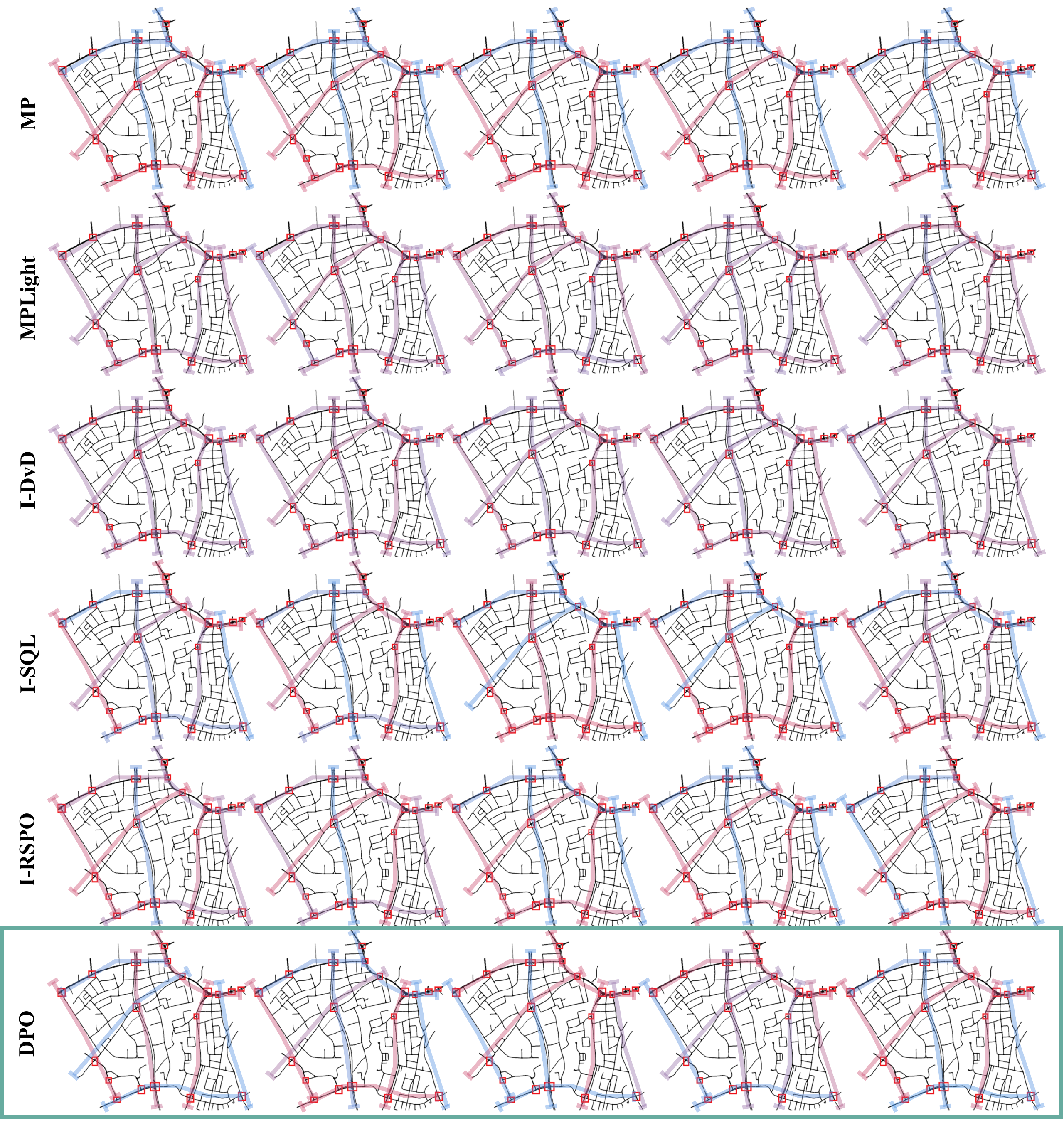}
        \caption{The Cologne scenario.}
        \label{fig:atsc-div-lng}
    \end{subfigure}
    \label{fig:atsc-div-all}
    \caption{Comparison of policy diversity among DPO and other baselines under two scenarios in ATSC benchmark. Different colors represent different commute times.}
\end{figure}

\begin{figure}[htb!]
    \centering
    \begin{subfigure}[b]{0.48\textwidth}
        \centering
        \includegraphics[width=\textwidth]{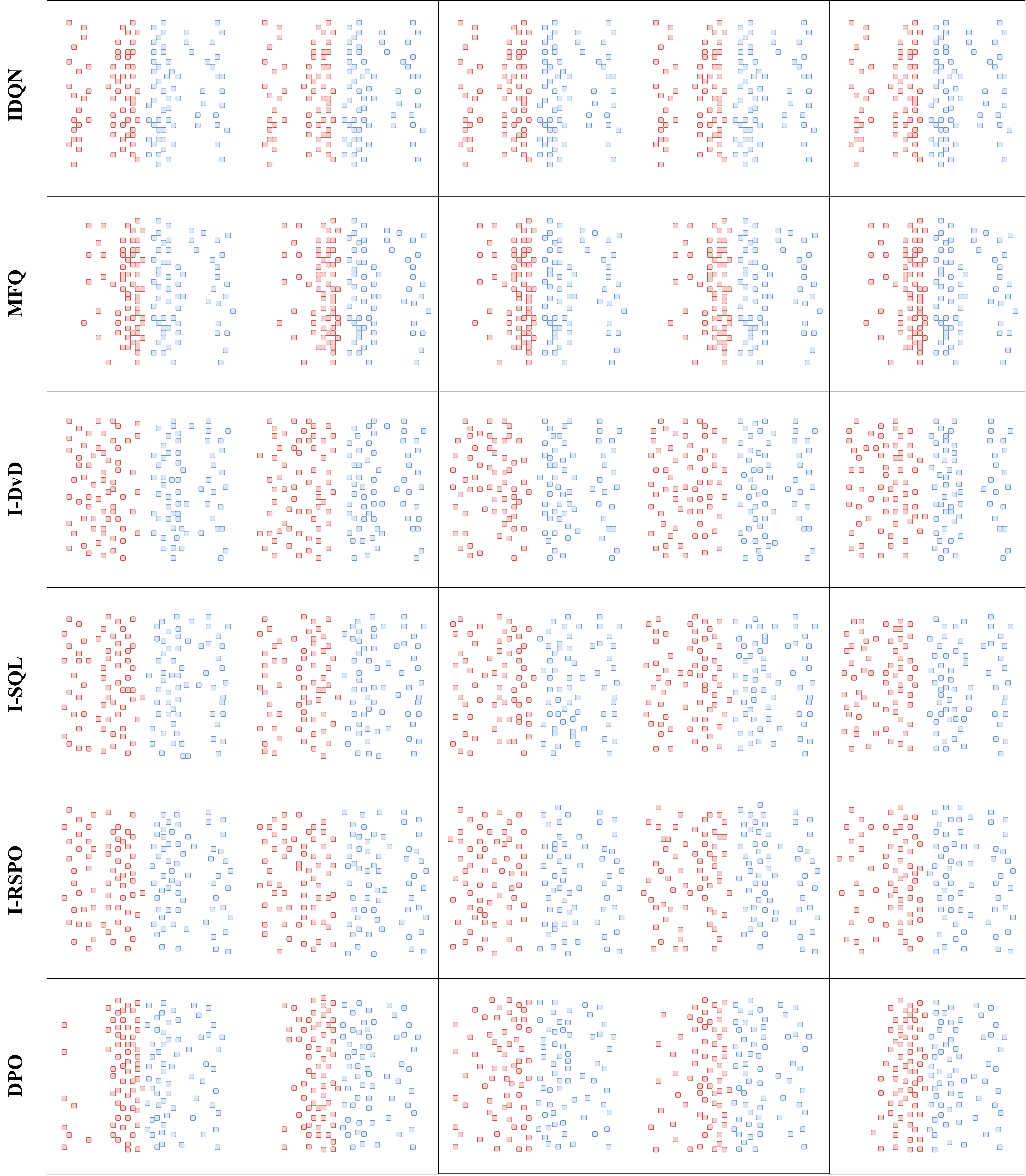}
        \caption{The early stage.}
        \label{fig:battle-div-begin}
    \end{subfigure}
    \begin{subfigure}[b]{0.48\textwidth}
        \centering
        \includegraphics[width=\textwidth]{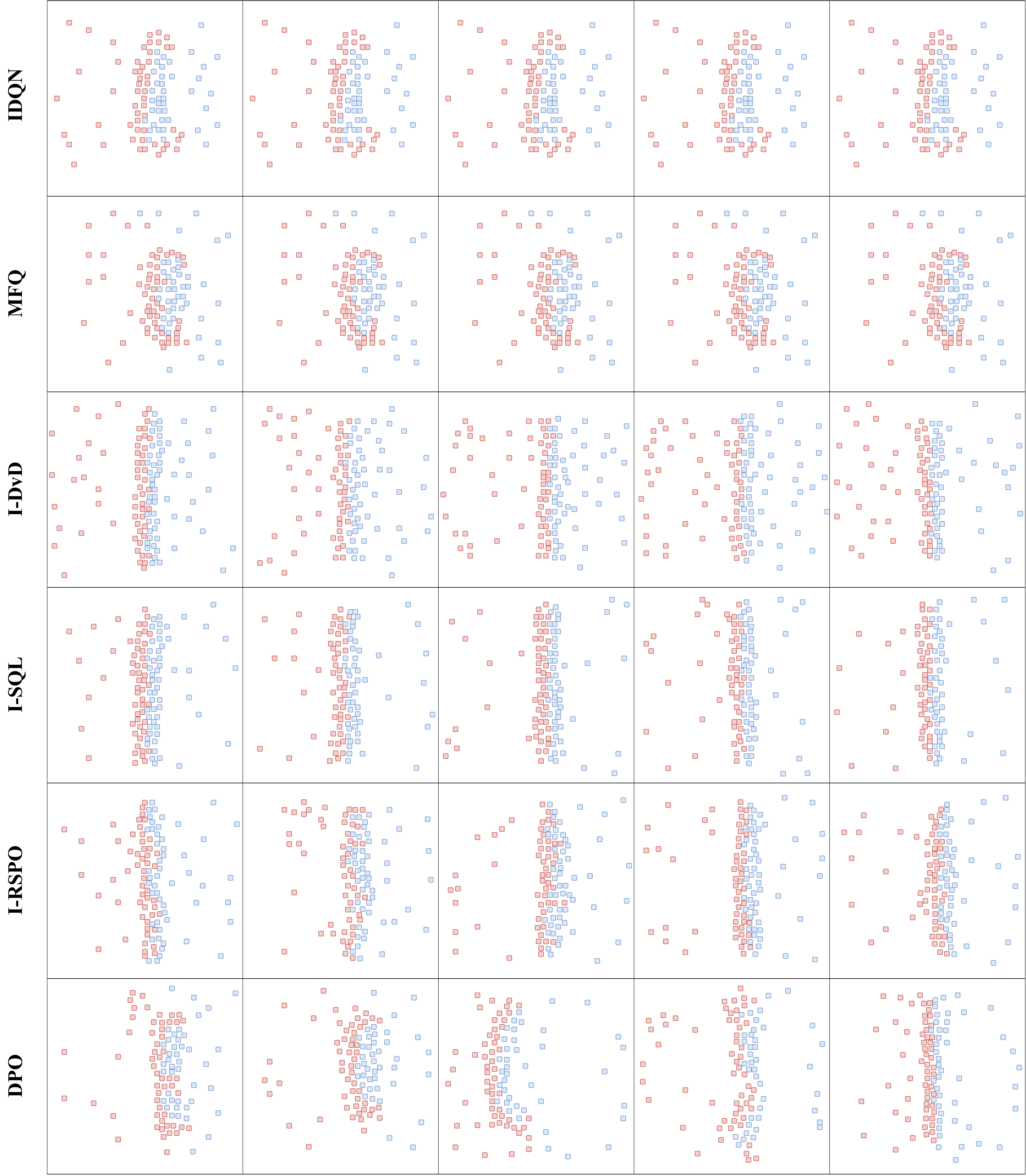}
        \caption{The middle stage.}
        \label{fig:battle-div-mid}
    \end{subfigure}
    \label{fig:battle-div}
    \caption{Comparison of policy diversity among DPO and others under the early and middle stages of the Battle benchmark.}
\end{figure}

Due to space constraints, we place some experimental results of the additional validation in the appendix section. 
These results consist of three main sections: one is a comparison of three algorithms that encourage policy diversity, DvD, SQL, and RSPO, with their respective independent learning variants; 
the second is ablation studies of the proposed DPO algorithm; 
Moreover, the third verify the robustness of different algorithms in the ATSC benchmark task under out-of-distribution traffic flow.
Before giving these additional experimental results, we post the complete diversity visualization results here, as shown in Figure~\ref{fig:atsc-div},~\ref{fig:atsc-div-lng},~\ref{fig:battle-div-begin} and~\ref{fig:battle-div-mid}.

% \begin{figure}[htb!]
%     \centering
%     \includegraphics[width=0.8\textwidth]{figs/atsc-diverse.pdf}
%     \caption{}
%     \label{fig:atsc-diverse}
% \end{figure}

% \begin{figure*}[htb!]
%     \centering
%     \includegraphics[width=0.8\textwidth]{figs/battle-div-begin.pdf}
%     \caption{}
%     \label{fig:battle-div-begin}
% \end{figure*}

% \begin{figure*}[htb!]
%     \centering
%     \includegraphics[width=0.8\textwidth]{figs/battle-div-mid.pdf}
%     \caption{}
%     \label{fig:battle-div-mid}
% \end{figure*}

\begin{figure}[htb!]
    \centering
    \begin{subfigure}[b]{0.45\textwidth}
        \centering
        \includegraphics[width=\textwidth]{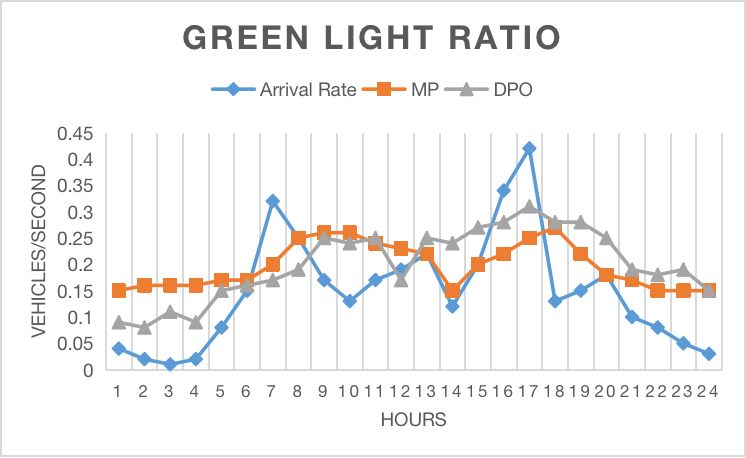}
        % \caption{The Ingostadt scenario.}
        \label{fig:green-ratio-1}
    \end{subfigure}
    \begin{subfigure}[b]{0.45\textwidth}
        \centering
        \includegraphics[width=\textwidth]{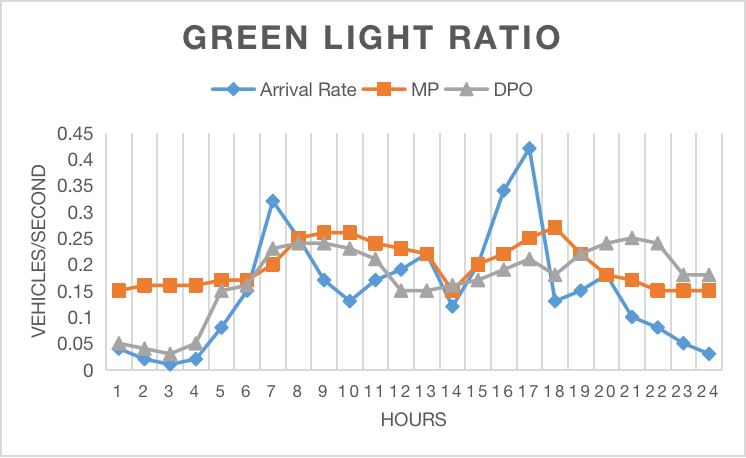}
        % \caption{The Ingostadt scenario.}
        \label{fig:green-ratio-2}
    \end{subfigure}
    \begin{subfigure}[b]{0.45\textwidth}
        \centering
        \includegraphics[width=\textwidth]{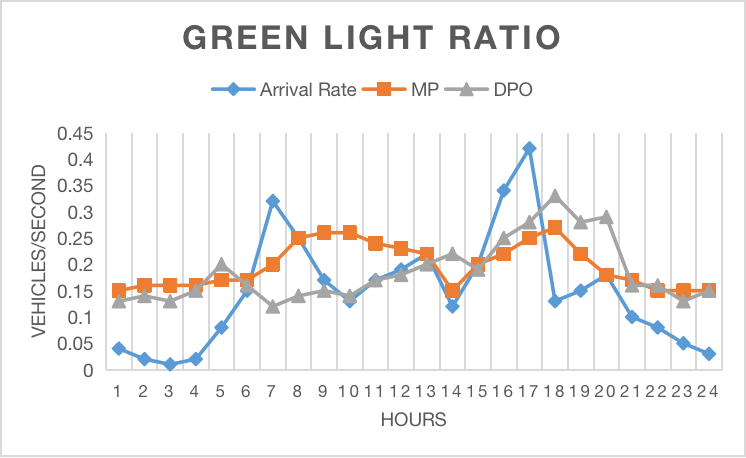}
        % \caption{The Ingostadt scenario.}
        \label{fig:green-ratio-3}
    \end{subfigure}
    \begin{subfigure}[b]{0.45\textwidth}
        \centering
        \includegraphics[width=\textwidth]{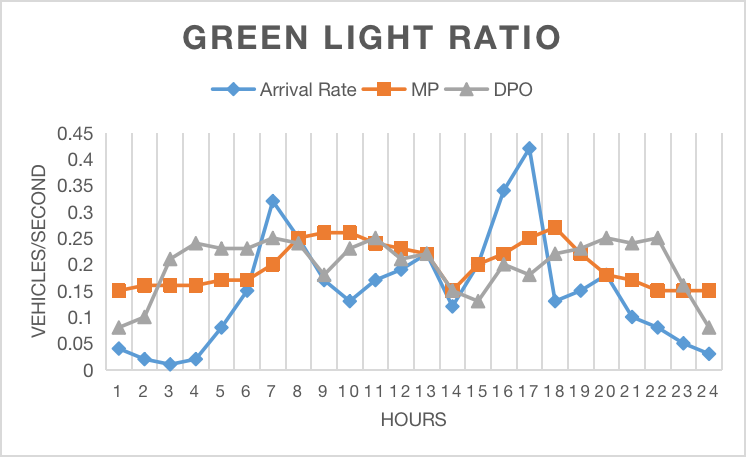}
        % \caption{The Ingostadt scenario.}
        \label{fig:green-ratio-4}
    \end{subfigure}
    \caption{The arrival rate of traffic in a specific direction at an intersection in a main road and the proportion of green lights generated by strategies output by different algorithms.}
    \label{fig:green-ratio}
\end{figure}

In addition to visualizing the global diversity of strategies obtained by different algorithms, we also show the proposed DPO's policy diversity at local intersections. 
In order to improve the interpretability of the visualization results, we selected the MP method based on heuristic rules and expert knowledge as a comparison, and the results are shown in Figure~\ref{fig:green-ratio}.
It can be seen from the figure that the strategy output by the MP can better match the traffic flow, thereby reducing the average delay and other indicators. 
However, the DPO method does not simply perform local optimization but considers global information. 
This makes the diversity policies obtained from the DPO achieve a trade-off for allocating green light time at different times of a single intersection.

\subsection{Independent Learning Variants}

In the ATSC benchmark task, we find that the performance of the three algorithms DvD, SQL, and RSPO, which encourage policy diversity, showed a significant degradation in large-scale structured action space. 
This is why in larger scale Battle scenarios, we directly use these algorithms' corresponding independent learning variants. 
In this section, we further compare the DvD, SQL, and RSPO algorithms and their independent learning variants I-DvD, I-SQL, and I-RSPO in the \textit{TAPAS Cologne} and \textit{InTAS} scenarios of the ATSC benchmark, and the results are shown in Table~\ref{tab:astc-ind-col} and~\ref{tab:astc-ind-ing}.

\begin{table}[htb!]
\caption{Performance ($\downarrow$) of independent learning variants on two scenarios of the ATSC benchmark.}
\begin{subtable}[t]{0.48\textwidth}
\centering
% \resizebox{0.48\columnwidth}{!}{%
\begin{tabular}{lll|lll}
\hline
\textit{\textbf{I-DvD}}.  & Col. Reg. & \textit{\textbf{DvD}}  & Col. Reg.      \\ \hline
Avg. Delay                & 50.16     & Avg. Delay             & 55.91          \\
Avg. Trip Time            & 108.43    & Avg. Trip Time         & 115.54         \\
Avg. Wait                 & 22.50     & Avg. Wait              & 28.35          \\
Avg. Queue                & 2.12      & Avg. Queue             & 2.28           \\ \hline
\textit{\textbf{I-SQL}}   & Col. Reg. & \textit{\textbf{SQL}}  & Col. Reg.      \\ \hline
Avg. Delay                & 28.39     & Avg. Delay             & 58.32          \\
Avg. Trip Time            & 93.48     & Avg. Trip Time         & 116.29         \\
Avg. Wait                 & 6.74      & Avg. Wait              & 30.01          \\
Avg. Queue                & 0.55      & Avg. Queue             & 2.06           \\ \hline
\textit{\textbf{I-RSPO}}  & Col. Reg. & \textit{\textbf{RSPO}} & Col. Reg.      \\ \hline
Avg. Delay                & 23.46     & Avg. Delay             & 57.28          \\
Avg. Trip Time            & 88.81     & Avg. Trip Time         & 120.53         \\
Avg. Wait                 & 5.95      & Avg. Wait              & 28.19          \\
Avg. Queue                & 0.49      & Avg. Queue             & 2.59          \\ \hline
\end{tabular}%
% }
\caption{Performance of independent learning variants (TAPAS Cologne).}
\label{tab:astc-ind-col}
\end{subtable}
\begin{subtable}[t]{0.48\textwidth}
\centering
\begin{tabular}{lll|lll}
\hline
\textit{\textbf{I-DvD}}.  & Ing. Reg. & \textit{\textbf{DvD}}  & Ing. Reg.      \\ \hline
Avg. Delay                & 74.58     & Avg. Delay             & 73.22          \\
Avg. Trip Time            & 215.07    & Avg. Trip Time         & 212.81         \\
Avg. Wait                 & 32.48     & Avg. Wait              & 31.36          \\
Avg. Queue                & 1.45      & Avg. Queue             & 1.42         \\ \hline
\textit{\textbf{I-SQL}}   & Ing. Reg. & \textit{\textbf{SQL}}  & Ing. Reg.      \\ \hline
Avg. Delay                & 65.29     & Avg. Delay             & 67.65          \\
Avg. Trip Time            & 201.26    & Avg. Trip Time         & 205.44         \\
Avg. Wait                 & 22.41     & Avg. Wait              & 26.45          \\
Avg. Queue                & 1.01      & Avg. Queue             & 1.15           \\ \hline
\textit{\textbf{I-RSPO}}  & Ing. Reg. & \textit{\textbf{RSPO}} & Ing. Reg.      \\ \hline
Avg. Delay                & 86.59     & Avg. Delay             & 90.42          \\
Avg. Trip Time            & 215.25    & Avg. Trip Time         & 226.5          \\
Avg. Wait                 & 39.76     & Avg. Wait              & 44.16          \\
Avg. Queue                & 1.58      & Avg. Queue             & 1.74           \\ \hline
\end{tabular}%
\caption{Performance of independent learning variants (InTAS).}
\label{tab:astc-ind-ing}
\end{subtable}
\label{tab:atsc-ind}

\end{table}

% \begin{table}[htb!]
% \resizebox{0.48\columnwidth}{!}{%
% \begin{tabular}{lll|lll}
% \hline
% \textit{\textbf{I-DvD}}.  & Ing. Reg. & \textit{\textbf{DvD}}  & Ing. Reg.      \\ \hline
% Avg. Delay                & 74.58     & Avg. Delay             & 73.22          \\
% Avg. Trip Time            & 215.07    & Avg. Trip Time         & 212.81         \\
% Avg. Wait                 & 32.48     & Avg. Wait              & 31.36          \\
% Avg. Queue                & 1.45      & Avg. Queue             & 1.42         \\ \hline
% \textit{\textbf{I-SQL}}   & Ing. Reg. & \textit{\textbf{SQL}}  & Ing. Reg.      \\ \hline
% Avg. Delay                & 65.29     & Avg. Delay             & 67.65          \\
% Avg. Trip Time            & 201.26    & Avg. Trip Time         & 205.44         \\
% Avg. Wait                 & 22.41     & Avg. Wait              & 26.45          \\
% Avg. Queue                & 1.01      & Avg. Queue             & 1.15           \\ \hline
% \textit{\textbf{I-RSPO}}  & Ing. Reg. & \textit{\textbf{RSPO}} & Ing. Reg.      \\ \hline
% Avg. Delay                & 86.59     & Avg. Delay             & 90.42          \\
% Avg. Trip Time            & 215.25    & Avg. Trip Time         & 226.5          \\
% Avg. Wait                 & 39.76     & Avg. Wait              & 44.16          \\
% Avg. Queue                & 1.58      & Avg. Queue             & 1.74           \\ \hline
% \end{tabular}%
% }
% \caption{Performance of independent learning variants on the InTAS scenario.}
% \label{tab:astc-ind-ing}
% \end{table}

\begin{figure}[htb!]
    \centering
    \begin{subfigure}[b]{0.53\textwidth}
        \centering
        \includegraphics[width=\textwidth]{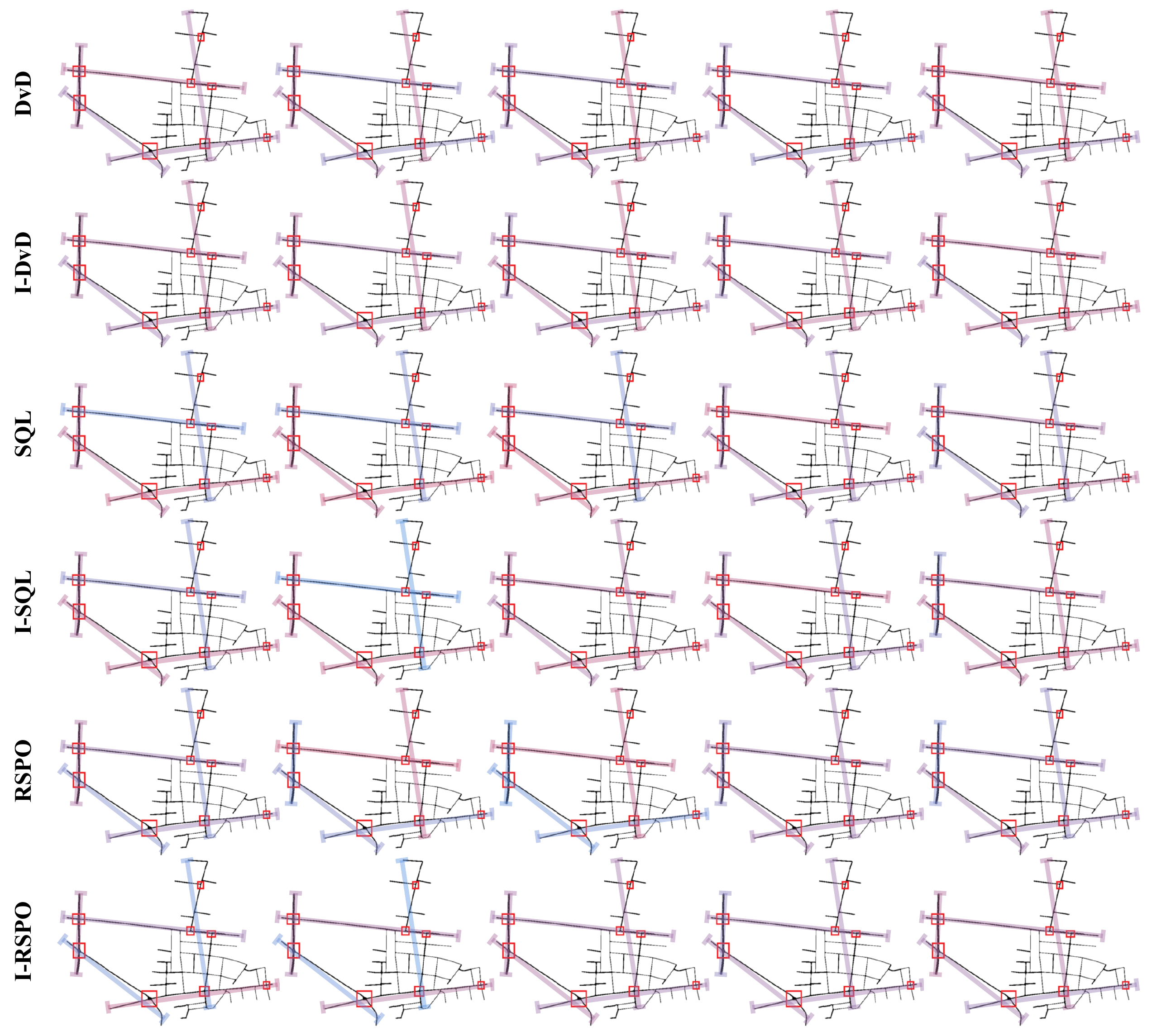}
        \caption{The Ingostadt scenario.}
        \label{fig:atsc-div-abl}
    \end{subfigure}
    \begin{subfigure}[b]{0.45\textwidth}
        \centering
        \includegraphics[width=\textwidth]{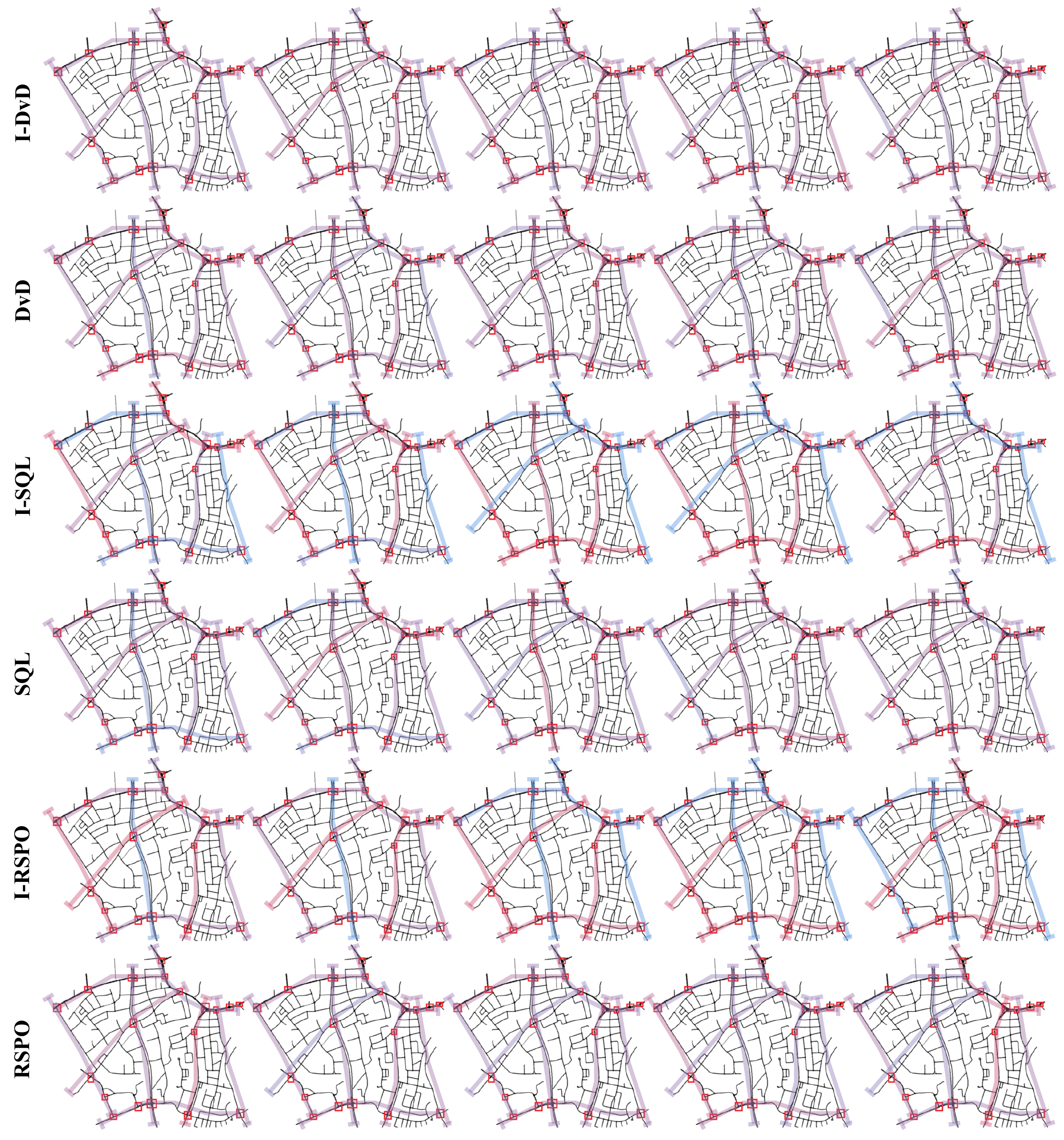}
        \caption{The Cologne scenario.}
        \label{fig:atsc-div-lng-abl}
    \end{subfigure}
    \caption{Comparison of policy diversity among baselines and their independent learning variants under two scenarios in ATSC benchmark. Different colors represent different commute times.}
    \label{fig:atsc-div-abl-all}
\end{figure}

The table shows that using the independent learning variant in a larger structured action space can lead to a more considerable performance improvement.
However, the DvD algorithm still does not perform as well as expected. 
Independent learning encourages diversity of atomic actions, which will also prevent I-SQL and I-RSPO from getting a better diversity of policies in the structured action space. 
To verify this, we used the same visualization method as in the experimental part of the main text, and the final results are shown in 
Figure~\ref{fig:atsc-div-abl-all}.
% Figure~\ref{fig:astc-div-abl} and~\ref{fig:astc-div-lng-abl}.

% \begin{figure*}[htb!]
%     \centering
%     \includegraphics[width=0.8\textwidth]{figs/atsc-div-abl.pdf}
%     \caption{}
%     \label{fig:astc-div-abl}
% \end{figure*}

% \begin{figure*}[htb!]
%     \centering
%     \includegraphics[width=0.8\textwidth]{figs/atsc-div-lng-abl.pdf}
%     \caption{}
%     \label{fig:astc-div-lng-abl}
% \end{figure*}

As can be seen from the figure, in the small-scale structured action space, the independent learning variant does not bring significant performance improvement in terms of diversity; 
However, in the large-scale structured action space, the independent learning variant learns policies with more significant diversity.

\subsection{Ablation Study}

In this section, we perform some ablation studies on the three critical implementations of the DPO algorithm, including the additional soft value regression (denoted as $\mathrm{S}$) task introduced to accelerate the training of total flow $Z$, the additionally expanded termination action (denoted as $\mathrm{T}$) to accelerate the training, and the action space design (denoted as $\mathrm{P}$) of GFlowNet. 
For the last point, in the ATSC benchmark, we analyze the impact of the road network-based GFlowNet's action space design on performance;
In the Battle benchmark, we analyze the impact of different physical topologies resulting from different influence ranges.

\begin{table}[htb!]
\caption{Ablation studies of the proposed DPO under two scenarios of the ATSC benchmark.}
\label{tab:atsc-ablation}
\resizebox{\columnwidth}{!}{%
\begin{tabular}{ccc|cccc|c|cccc|c}
\multirow{2}{*}{Soft value regression} &
  \multirow{2}{*}{Termination action} &
  \multirow{2}{*}{Physical dependencies} &
  \multicolumn{4}{c|}{Ing.Reg} &
  \multirow{2}{*}{Epochs} &
  \multicolumn{4}{c|}{Col.Reg} &
  \multirow{2}{*}{Epochs} \\ \cline{4-7} \cline{9-12}
             &              &              & Delay & Trip   & Wait  & Queue  &                   & Delay & Trip   & Wait  & Queue &                   \\ \hline
             &              &              & 78.92 & 214.63 & 32.75 & 1.51   & /                & 57.6  & 120.85 & 31.66 & 2.43 & /                \\
$\checkmark$ &              &              & 59.79 & 198.85 & 18.67 & 0.71   & $\sim$2.6$\times$ & 23.23 & 85.2   & 4.88  & 0.33  & $\sim$3.5$\times$ \\
             & $\checkmark$ &              & 77.06 & 210.49 & 29.68 & 1.48   & /                & 61.53 & 125.12 & 33.75 & 2.64  & /                \\
             &              & $\checkmark$ & 78.61 & 218.42 & 32.89 & 1.52   & /                & 60.99 & 126.4  & 33.71 & 2.61 & /                \\
$\checkmark$ & $\checkmark$ &              & 72.4  & 200.72 & 23.45 & 1.39   & $\sim$2$\times$   & 30.26 & 91.59  & 8.81  & 0.62  & $\sim$1.7$\times$ \\
             & $\checkmark$ & $\checkmark$ & 78.5  & 211.46 & 32.57 & 1.51   & /                & 58.76 & 123.58 & 31.8  & 2.54 & /                \\
$\checkmark$ &
   &
  $\checkmark$ &
  59.35 &
  \textbf{194.16} &
  \textbf{18.23} &
  \textbf{0.65} &
  $\sim$2.6$\times$ &
  \textbf{20.22} &
  85.49 &
  4.86 &
  0.33 &
  $\sim$3.2$\times$ \\
$\checkmark$ &
  $\checkmark$ &
  $\checkmark$ &
  \textbf{57.2} &
  192.75 &
  18.26 &
  \textbf{0.65} &
  \textbf{1$\times$} &
  \textbf{20.28} &
  \textbf{81.42} &
  \textbf{4.77} &
  \textbf{0.32} &
  \textbf{1$\times$}
\end{tabular}%
}

\end{table}

\begin{figure}[htb!]
    \centering
    \begin{subfigure}[b]{0.23\textwidth}
        \centering
        \includegraphics[width=\textwidth]{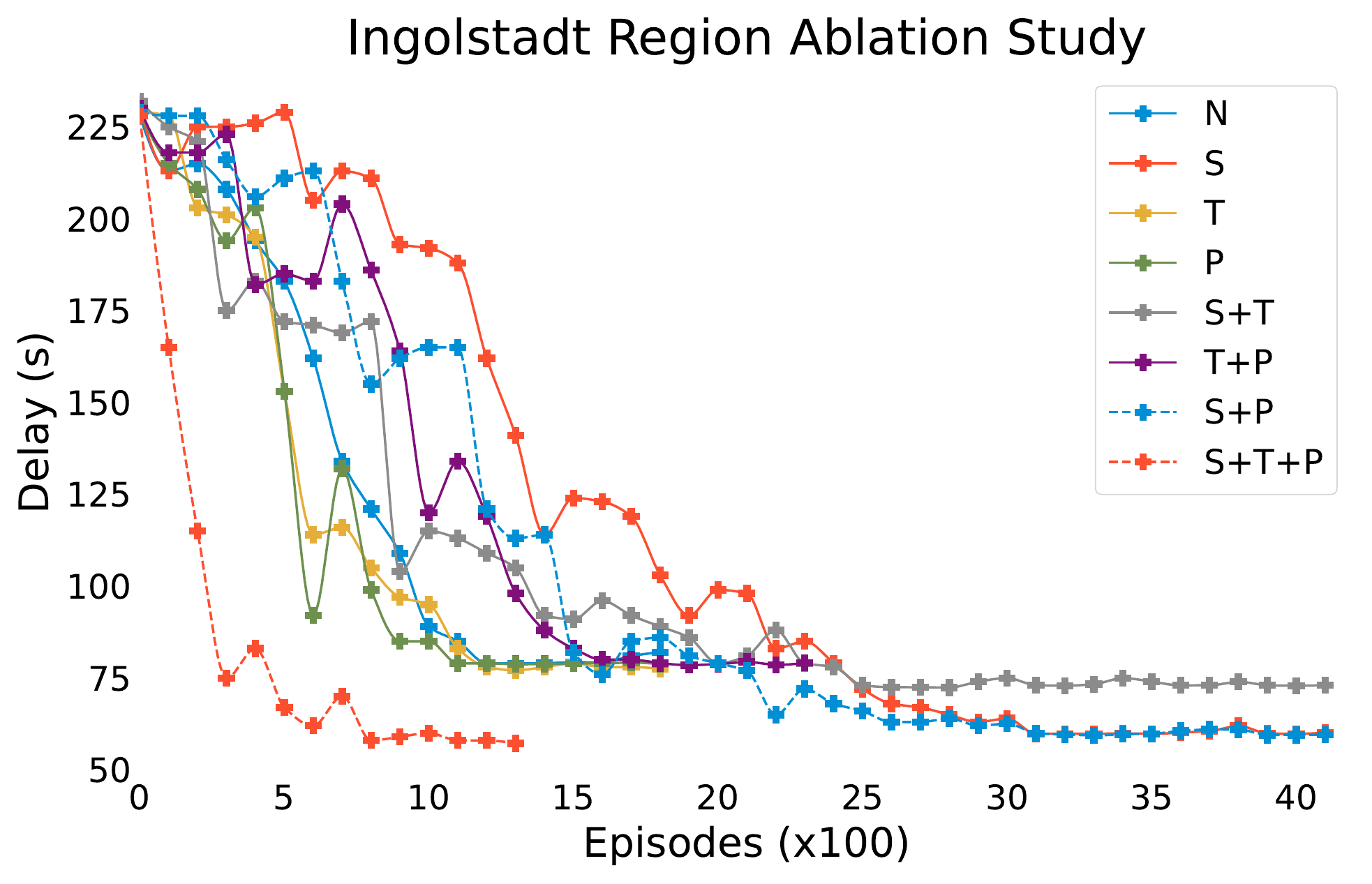}
        \caption{The Ingostadt scenario.}
        \label{fig:atsc-abl-ing}
    \end{subfigure}
    \begin{subfigure}[b]{0.23\textwidth}
        \centering
        \includegraphics[width=\textwidth]{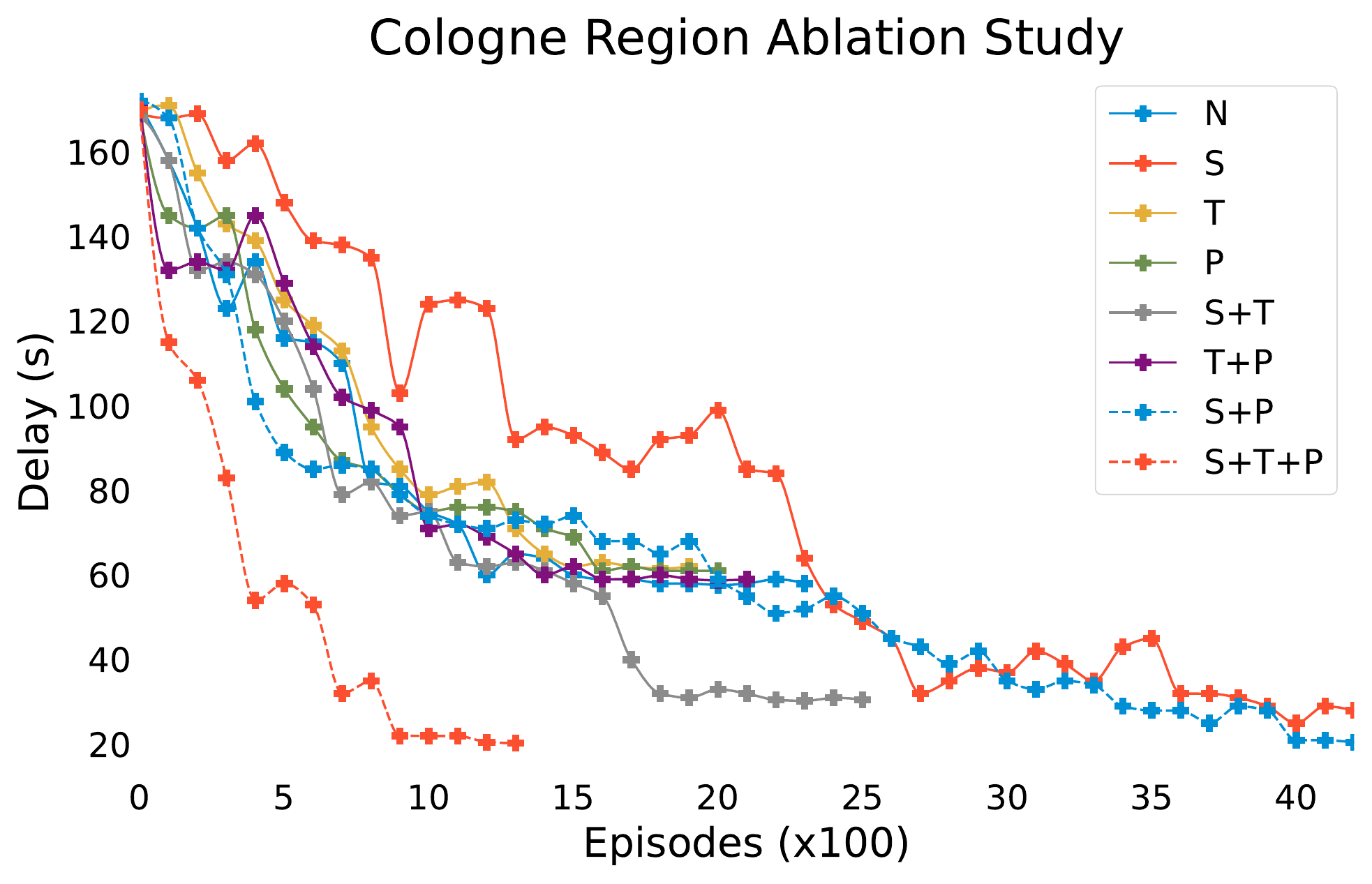}
        \caption{The Cologne scenario.}
        \label{fig:atsc-abl-col}
    \end{subfigure}
    \begin{subfigure}[b]{0.23\textwidth}
        \centering
        \includegraphics[width=\textwidth]{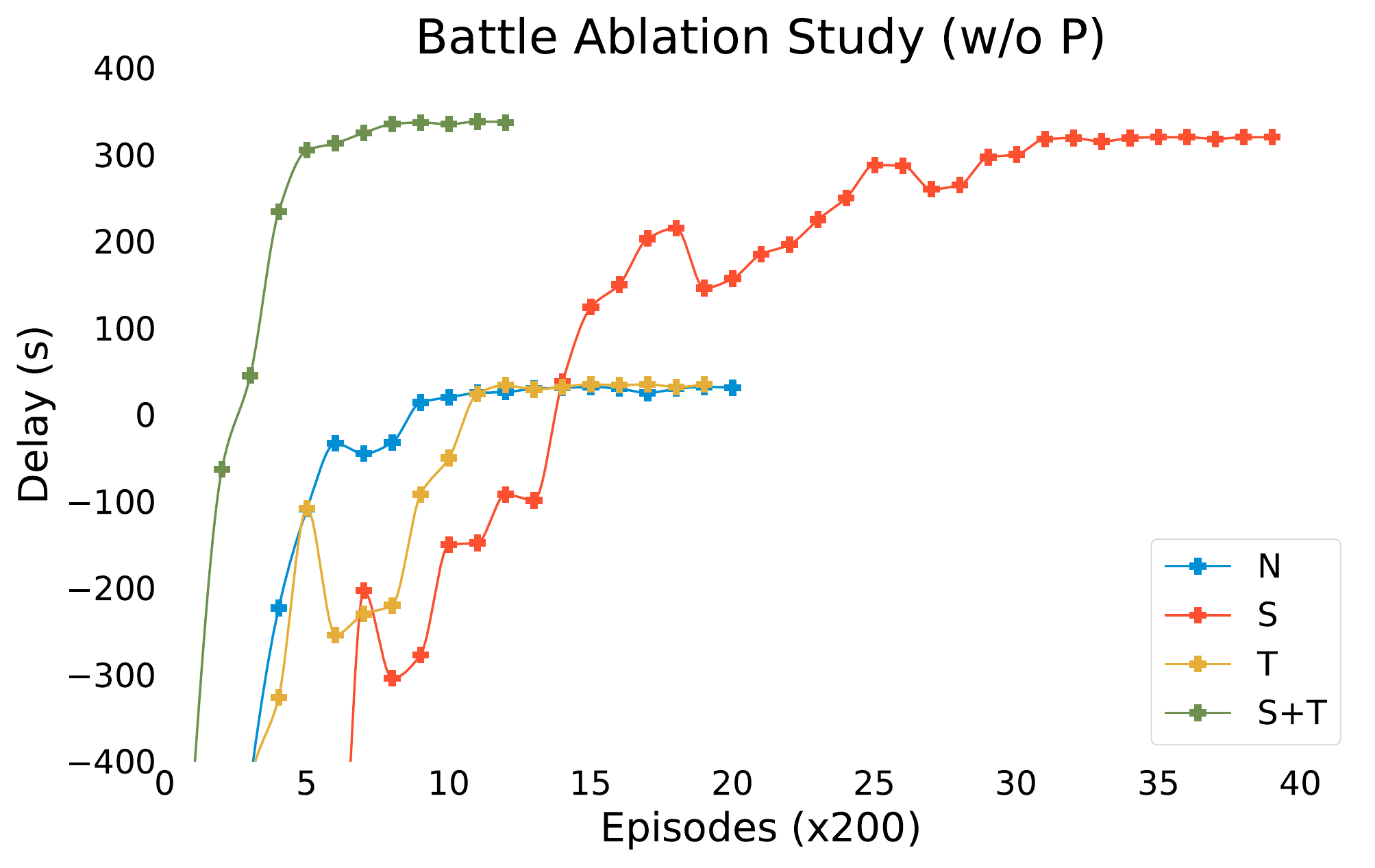}
        \caption{The Battle scenario.}
        \label{fig:battle-abl-1}
    \end{subfigure}
    \begin{subfigure}[b]{0.23\textwidth}
        \centering
        \includegraphics[width=\textwidth]{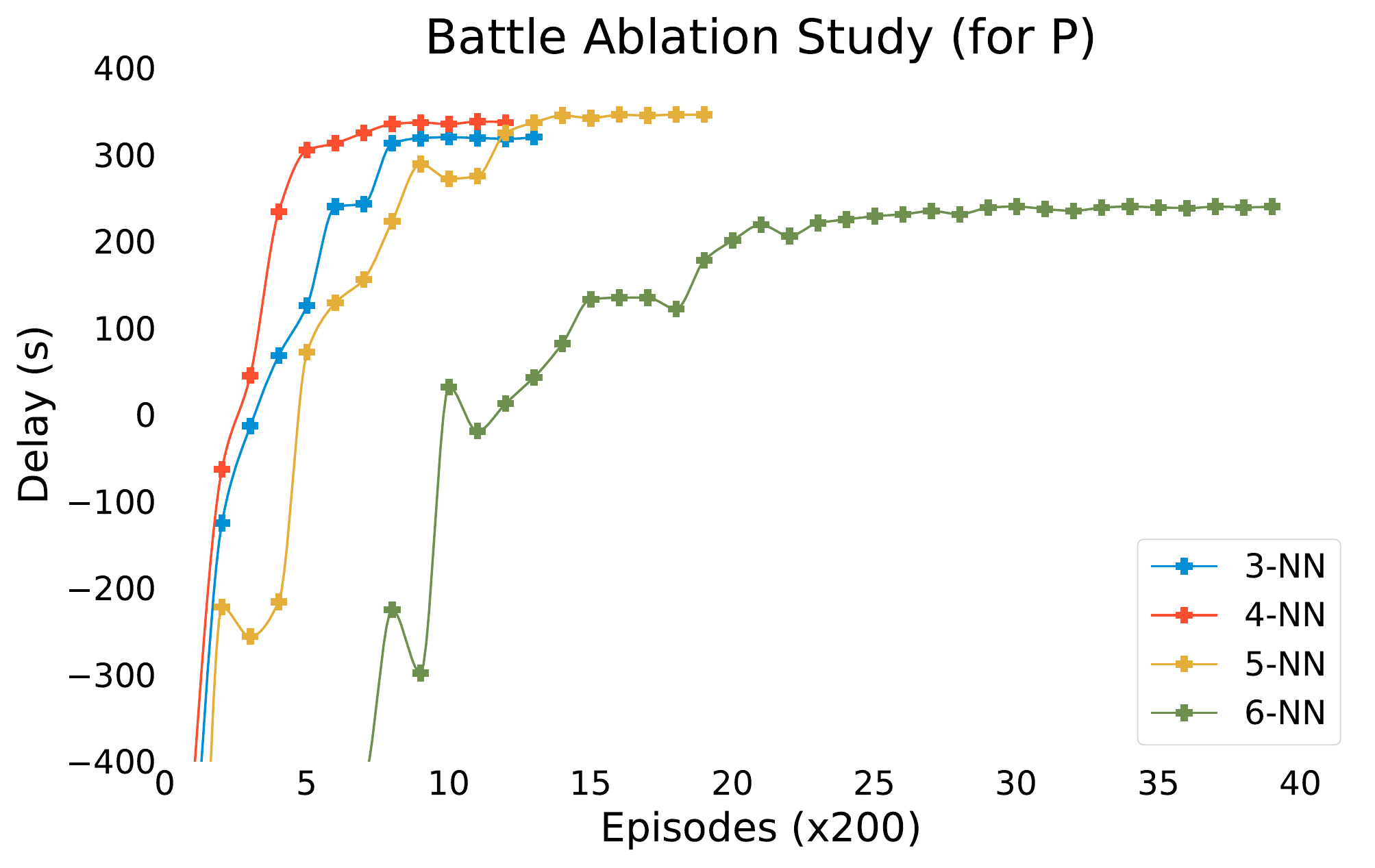}
        \caption{The Battle scenario.}
        \label{fig:battle-abl-2}
    \end{subfigure}
    \label{fig:curve-ablation}
    \caption{Ablation studies on the ATSC and Battle benchmarks, where $\mathrm{S}$ denotes the additional soft value regression task introduced to accelerate the training of total flow $Z$, $\mathrm{T}$) denotes the additionally expanded termination action accelerate the training, and $\mathrm{P}$ denotes the action space design of the GFlowNet. $N$ denotes algorithms without above three techaniques.}
\end{figure}

% \begin{figure}[htb!]
%     \centering
%     \begin{subfigure}[b]{0.45\textwidth}
%         \centering
%         \includegraphics[width=\textwidth]{figs/Battle-abl-1.pdf}
%         \caption{}
%     \end{subfigure}
%     \begin{subfigure}[b]{0.45\textwidth}
%         \centering
%         \includegraphics[width=\textwidth]{figs/Battle-abl-2.pdf}
%         \caption{}
%     \end{subfigure}
%     \label{fig:my_label}
% \end{figure}

We first analyze the performance of the DPO algorithm on the ATSC benchmark, and the results are shown in Table~\ref{tab:atsc-ablation}, Figure~\ref{fig:atsc-abl-ing} and~\ref{fig:atsc-abl-col}. 
As seen from the table, the soft value regression task plays a crucial role in the performance of the DPO. 
This is due to its operational guidance for training total flow $Z$, and the accuracy of $Z$ estimation directly determines the diversity of the sampled structured actions.
While the termination action and the road network-based GFlowNet's action space design have little impact, they can significantly improve the convergence speed of the algorithm. 
Overall, the results of the ablation study are consistent with our previous conjecture.

\begin{table}[htb!]
\caption{Ablation studies of the proposed DPO under the Battle benchmark.}
\label{tab:battle-ablation}
\begin{subtable}[t]{0.48\textwidth}
\resizebox{\columnwidth}{!}{%
\centering
\begin{tabular}{cc|cc|c}
Soft value regression & Termination action & Avg. \# Kills             & Avg. \# Reward              & Avg. Epochs        \\ \hline
                      &                    & 20.1($\pm 3.9$)          & 0.013($\pm 0.02$)          & /        \\
$\checkmark$          &                    & 61.3($\pm 0.4$)          & 0.135($\pm 0.08$)          & $\sim$3.7$\times$        \\
                      & $\checkmark$       & 20.6($\pm 4.2$)          & 0.015($\pm 0.02$)          & /                  \\
$\checkmark$          & $\checkmark$       & \textbf{62.1($\pm$ 0.1)} & \textbf{0.142($\pm$ 0.19)} & \textbf{1$\times$}
\end{tabular}%
}
% \caption{}
\label{tab:battle-ablation-1}
\end{subtable}
\begin{subtable}[t]{0.48\textwidth}
\begin{tabular}{c|cc|c}
\# Nearest agents & Avg. \# Kills             & Avg. \# Reward              & Avg. Epochs              \\ \hline
3                & 61.3($\pm 0.4$)          & 0.135($\pm 0.08$)          & \textbf{$\sim$1$\times$} \\
4                & \textbf{62.1($\pm$ 0.1)} & \textbf{0.142($\pm$ 0.19)} & \textbf{1$\times$}       \\
5                & \textbf{62.3($\pm$ 0.1)} & \textbf{0.146($\pm$ 0.23)} & $\sim$1.7$\times$        \\
6                & 58.6($\pm$ 0.1)          & 0.101($\pm$ 0.25)          & $\sim$3.1$\times$       
\end{tabular}%
% \caption{}
\label{tab:battle-ablation-2}
\end{subtable}

\end{table}

% \begin{table}[htb!]
% \resizebox{0.5\columnwidth}{!}{%
% \begin{tabular}{c|cc|c}
% \# Nearest agents & Avg. \# Kills             & Avg. \# Reward              & Avg. Epochs              \\ \hline
% 3                & 61.3($\pm 0.4$)          & 0.135($\pm 0.08$)          & \textbf{$\sim$1$\times$} \\
% 4                & \textbf{62.1($\pm$ 0.1)} & \textbf{0.142($\pm$ 0.19)} & \textbf{1$\times$}       \\
% 5                & \textbf{62.3($\pm$ 0.1)} & \textbf{0.146($\pm$ 0.23)} & $\sim$1.7$\times$        \\
% 6                & 58.6($\pm$ 0.1)          & 0.101($\pm$ 0.25)          & $\sim$3.1$\times$       
% \end{tabular}%
% }
% \caption{}
% \label{tab:battle-ablation-2}
% \end{table}

The ablation studies on the DPO algorithm in the Battle benchmark task exhibited similar results, as shown in Table~\ref{tab:battle-ablation}, Figure~\ref{fig:battle-abl-1} and~\ref{fig:battle-abl-2}.
In our experiments, instead of picking a different range of influence, an alternative approach is used, i.e., the nearest $k$ agents are chosen for implementation. 
As seen in Table~\ref{tab:battle-ablation}, while choosing a more significant number of agents to form the physical dependencies provides a slight performance improvement, it also slows down the convergence of the algorithm because of the resulting larger GFlowNet action space.

\subsection{Robustness in ATSC benchmarks}

As explained in the $\S$\ref{sec:intro}, diversity of policies can improve the robustness of algorithms in non-stationary environments. 
Therefore, this section tests the robustness of different algorithms by perturbing the traffic distribution in the ATSC benchmark and verifies whether the diversity policies are effective against the non-stationary factors in the environment.
Specifically, for the TAPAS Cologne ($8$ lights, $5$ main roads) and InTAS ($21$ lights, $8$ main roads) scenarios in the ATSC benchmark, we first randomly select $1$ or $2$ of the respective main roads, increase the traffic flow by $10\%$, and train all the algorithms for $50$ episodes (about $3\%$ of the standard training sample size). 

Since the DvD and MPLight algorithms have poor performance under ATSC and Battle benchmarks, we do not consider these two methods here. 
Also, considering the poor scalability of SQL and RSPO under large-scale structured action spaces, we only verify the robustness of the independent learning variants, i.e., I-SQL and I-RSPO.
The average performance is shown in Table~\ref{tab:robust}.

\begin{table}[htb!]
\caption{The robustness of different algorithms after perturbing the traffic distribution in the ATSC benchmark. we randomly select $1$ or $2$ of the respective main roads, increase the traffic flow by $10\%$, and traine all the algorithms for $50$ episodes (about $3\%$ of the standard training sample size).}
\label{tab:robust}
\resizebox{0.8\columnwidth}{!}{%
\begin{tabular}{l|ll|ll|l|ll|ll}
\hline
\multicolumn{1}{c|}{\multirow{2}{*}{\textit{\textbf{MP}}}} & \multicolumn{2}{c|}{Ing. Reg.} & \multicolumn{2}{c|}{Col. Reg.} & \multicolumn{1}{c|}{\multirow{2}{*}{\textit{\textbf{I-SQL}}}} & \multicolumn{2}{c|}{Ing. Reg.} & \multicolumn{2}{c}{Col. Reg.} \\ \cline{2-5} \cline{7-10} 
\multicolumn{1}{c|}{} & \multicolumn{1}{c}{1 road} & \multicolumn{1}{c|}{2 roads} & \multicolumn{1}{c}{1 road} & \multicolumn{1}{c|}{2 roads} & \multicolumn{1}{c|}{} & \multicolumn{1}{c}{1 road} & \multicolumn{1}{c|}{2 roads} & \multicolumn{1}{c}{1 road} & \multicolumn{1}{c}{2 roads} \\ \hline
Avg. Delay & 70.24 & 85.9 & 26.83 & 31.4 & Avg. Delay & 91.36 & 101.04 & 38.31 & 44.59 \\
Avg. Trip Time & 235.77 & 272.82 & 92.02 & 113.25 & Avg. Trip Time & 258.09 & 286.68 & 134.58 & 143.06 \\
Avg. Wait & 23.46 & 26.18 & 5.59 & 6.31 & Avg. Wait & 32.31 & 35.73 & 7.36 & 7.87 \\
Avg. Queue & 0.83 & 0.88 & 0.39 & 0.43 & Avg. Queue & 1.26 & 1.39 & 0.52 & 0.59 \\ \hline
\multicolumn{1}{c|}{\multirow{2}{*}{\textit{\textbf{I-RSPO}}}} & \multicolumn{2}{c|}{Ing. Reg.} & \multicolumn{2}{c|}{Col. Reg.} & \multicolumn{1}{c|}{\multirow{2}{*}{\textit{\textbf{DPO}}}} & \multicolumn{2}{c|}{Ing. Reg.} & \multicolumn{2}{c}{Col. Reg.} \\ \cline{2-5} \cline{7-10} 
\multicolumn{1}{c|}{} & 1 road & 2 roads & 1 road & 2 roads & \multicolumn{1}{c|}{} & 1 road & 2 roads & 1 road & 2 roads \\ \hline
Avg. Delay & 88.54 & 97.46 & 35.51 & 39.62 & Avg. Delay & \textbf{60.27} & \textbf{72.77} & \textbf{21.45} & \textbf{25.37} \\
Avg. Trip Time & 246.73 & 261.9 & 126.99 & 132.55 & Avg. Trip Time & \textbf{211.82} & \textbf{245.41} & \textbf{86.6} & \textbf{102.87} \\
Avg. Wait & 30.63 & 32.23 & 7.08 & 7.48 & Avg. Wait & \textbf{19.17} & \textbf{22.91} & \textbf{5.04} & \textbf{5.87} \\
Avg. Queue & 1.15 & 1.22 & 0.47 & 0.51 & Avg. Queue & \textbf{0.69} & \textbf{0.824} & \textbf{0.34} & \textbf{0.42} \\ \hline
\end{tabular}%
}

\end{table}

As seen from the table, DPO can quickly achieve good performance using only a small number of samples for fine-tuning. 
The lack of policy diversity in the other algorithms makes them have a significant performance gap with DPO.

% %%% The following command should be issued somewhere in the first column 
% %%% of the final page of your paper.
% \balance

%%%%%%%%%%%%%%%%%%%%%%%%%%%%%%%%%%%%%%%%%%%%%%%%%%%%%%%%%%%%%%%%%%%%%%%%

\end{document}